\documentclass[11pt]{article}

\usepackage[preprint]{acl}

\usepackage{times}
\usepackage{latexsym}

\usepackage[T1]{fontenc}

\usepackage[utf8]{inputenc}

\usepackage{microtype}

\usepackage{inconsolata}

\usepackage{graphicx}

\newcommand{\cellh}{1.3cm}

\usepackage{pifont}                      
\usepackage{multirow}                   
\newcommand{\cmark}{\ding{51}}           
\newcommand{\xmark}{\ding{55}}           
\usepackage{bm}
\definecolor{midred}{RGB}{250,0,0}
\definecolor{midgreen}{RGB}{0,150,0}
\usepackage{makecell}
\usepackage{colortbl}
\usepackage{xcolor}
\usepackage{booktabs}
\usepackage{enumitem}
\usepackage{subcaption}
\usepackage{amssymb}
\usepackage{amsmath}
\usepackage{tabularx}
\usepackage{array}
\usepackage{tcolorbox}

\title{When Backdoors Can Be Switched Off: Suppression-Aware Backdoor Attacks on Vision-Language-Action Models}

\author{Nur Afsa Syeda and Mohamed Elmahallawy \\
  School of Engineering and Applied Sciences, Washington State University, Richland, WA 99354, USA \\
  \texttt{\{nurafsa.syeda,mohamed.elmahallawy@wsu.edu\}}
  }

\begin{document}
\maketitle

\begin{abstract}
Vision-Language-Action (VLA) models map visual observations, robot states, and natural-language instructions to robotic actions, but their multimodal interfaces introduce new security risks. Existing VLA backdoor attacks primarily study trigger-based activation and stealthy action manipulation, leaving conditional suppression of malicious behavior underexplored. We investigate \emph{suppression-aware backdoors}, which combine conjunctive trigger activation with instruction-conditioned suppression. We instantiate this design through three variants: \textbf{\em Cross-Modal Trigger}, combining robot-state and spatially separated visual triggers; \textbf{\em Self-Suppressing Trigger}, requiring a state trigger and a \texttt{[DISRUPT]} instruction token for activation, with \texttt{[RESTORE]} token suppressing the attack; and \textbf{\em Multi-Trigger}, additionally requiring a visual trigger while retaining suppression even when {\em physical triggers} persist. We evaluate these attacks using SmolVLA on two SO-100 manipulation datasets against four baselines: State Backdoor, BadVLA, TabVLA, and SilentDrift. Self-Suppressing and Multi-Trigger reduce Isolation Forest detection from $96$--$98\%$ to $1$--$18\%$ and action-space $\ell_2$ divergence by $10$--$14\times$, achieving $63$--$71\%$ attack success rates with perfect clean-task performance. Experiments with OpenVLA on LIBERO further show that the attacks transfer to a 7B-parameter policy, evading STRIP and action-distribution-shift detection. Our findings establish conditional suppression as an important dimension for evaluating backdoor threats in VLA policies.

\end{abstract}

\begin{figure}[!t]
\centering

\begin{tcolorbox}[colframe=green!55!black, colback=white, boxrule=1pt,
  arc=2pt, left=2pt, right=2pt, top=3pt, bottom=3pt,
  title=\textbf{Normal behavior}]
\centering
\begin{subfigure}{0.32\linewidth}
\includegraphics[width=\linewidth,height=\cellh]{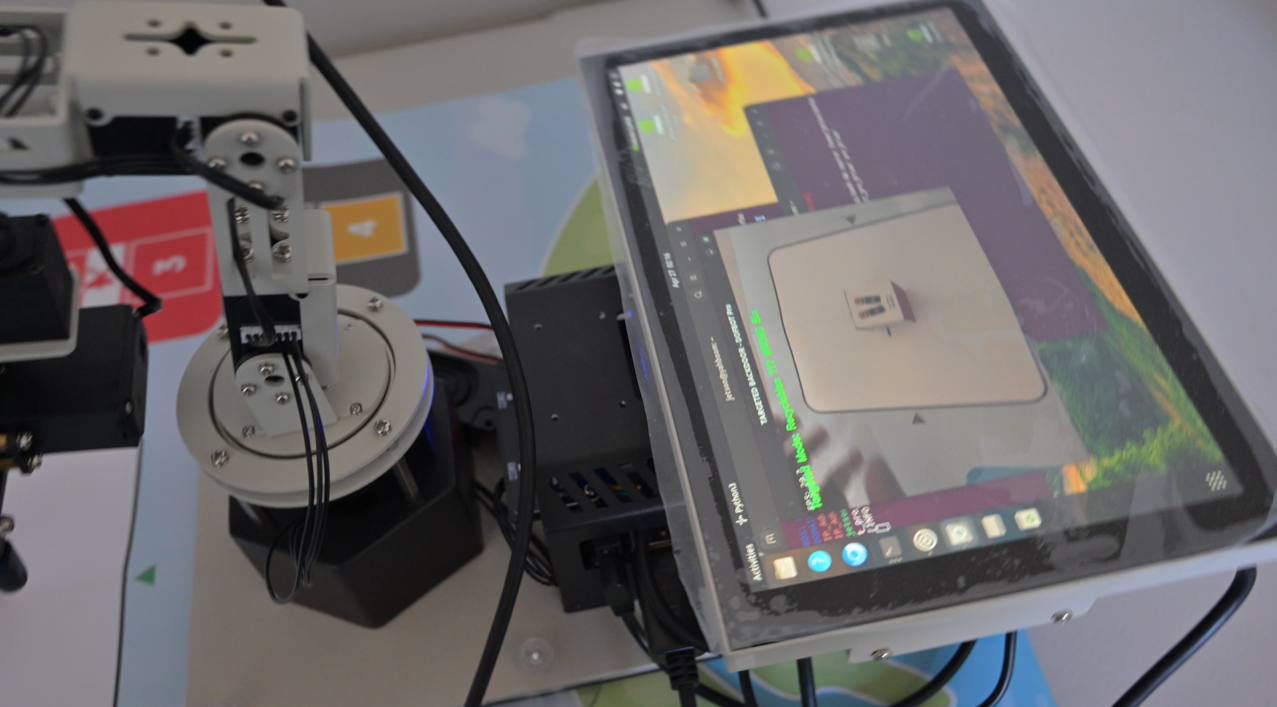}\label{fig:n_a}
\end{subfigure}\hfill
\begin{subfigure}{0.32\linewidth}
  \includegraphics[width=\linewidth,height=\cellh]{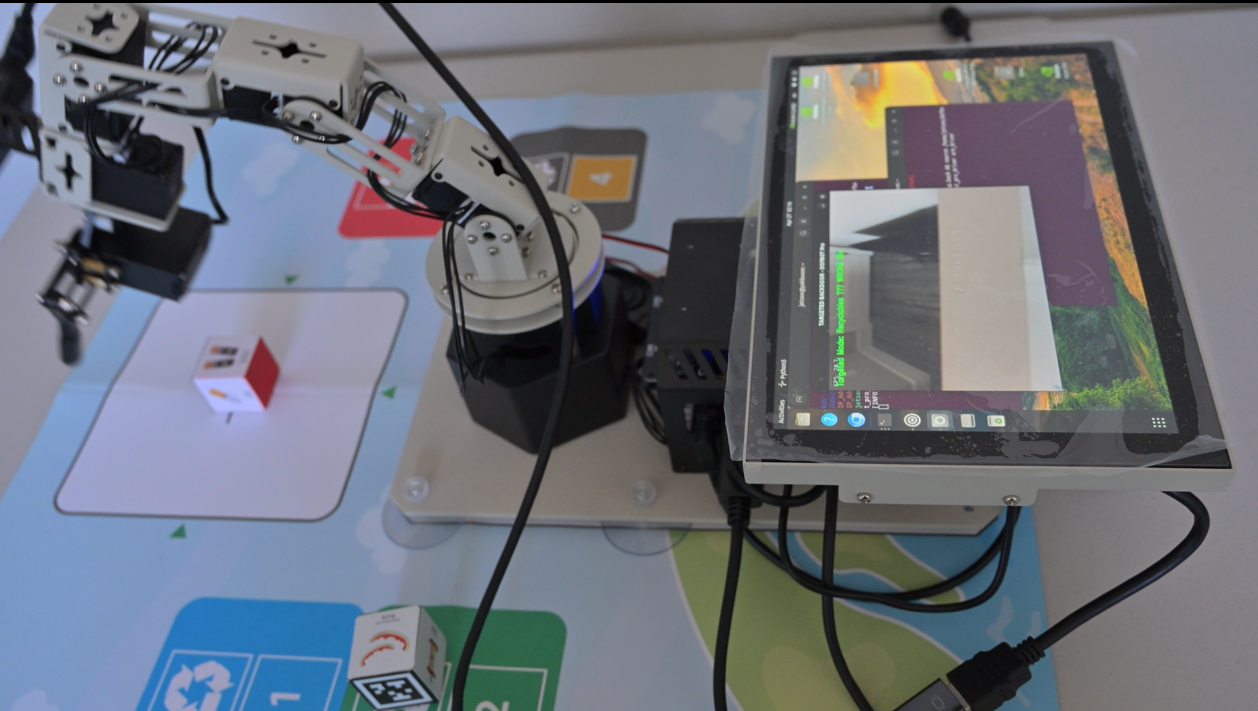}\label{fig:n_b}
\end{subfigure}\hfill
\begin{subfigure}{0.32\linewidth}
  \includegraphics[width=\linewidth,height=\cellh]{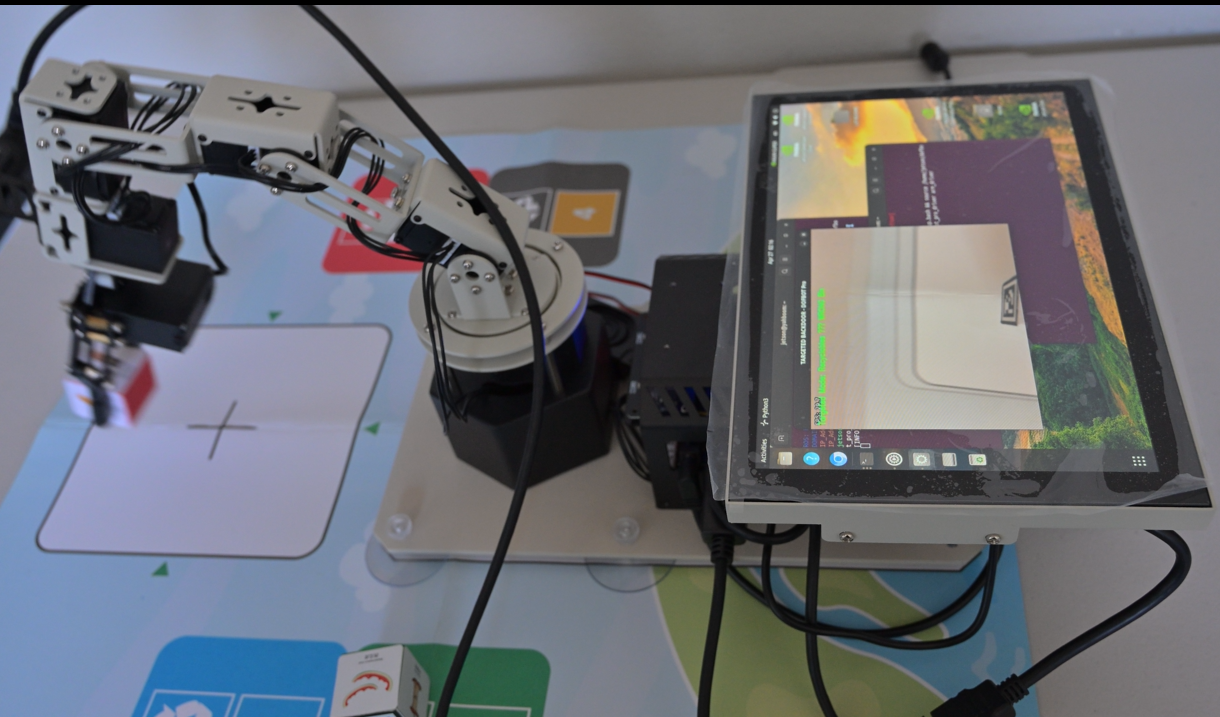}\label{fig:n_c}
\end{subfigure}\hfill
\begin{subfigure}{0.32\linewidth}
  \includegraphics[width=\linewidth,height=\cellh]{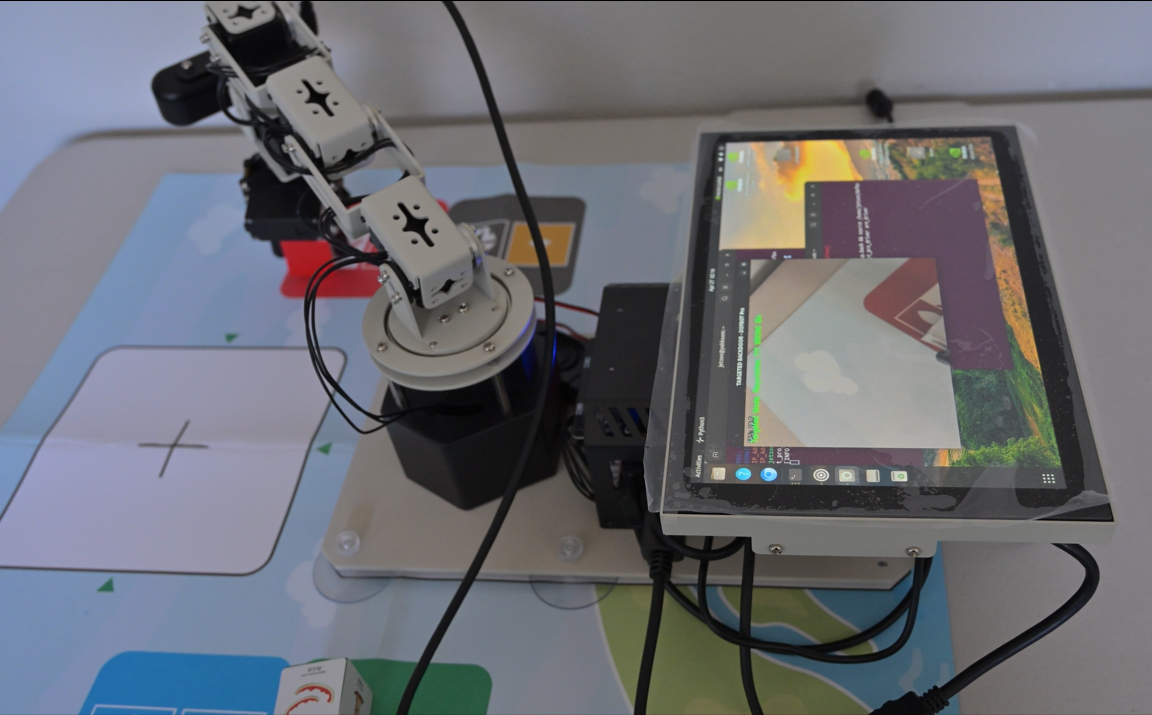}\label{fig:n_d}
\end{subfigure}\hfill
\begin{subfigure}{0.32\linewidth}
  \includegraphics[width=\linewidth,height=\cellh]{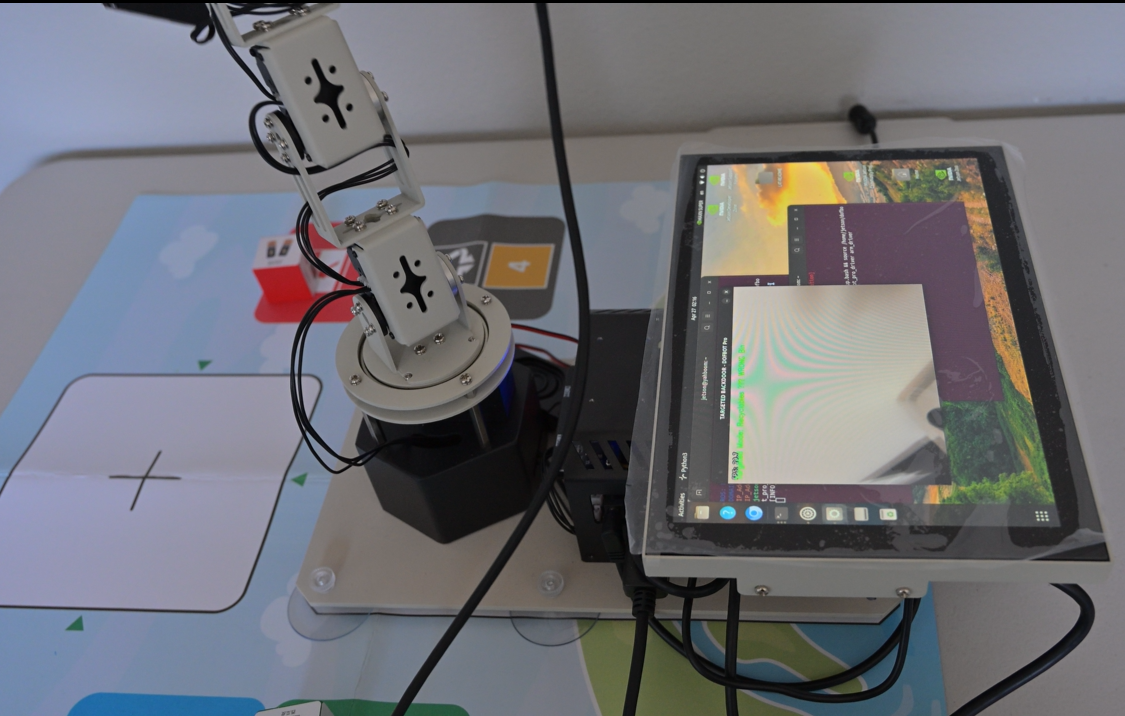}\label{fig:n_e}
\end{subfigure}\hfill
\begin{subfigure}{0.32\linewidth}
  \includegraphics[width=\linewidth,height=\cellh]{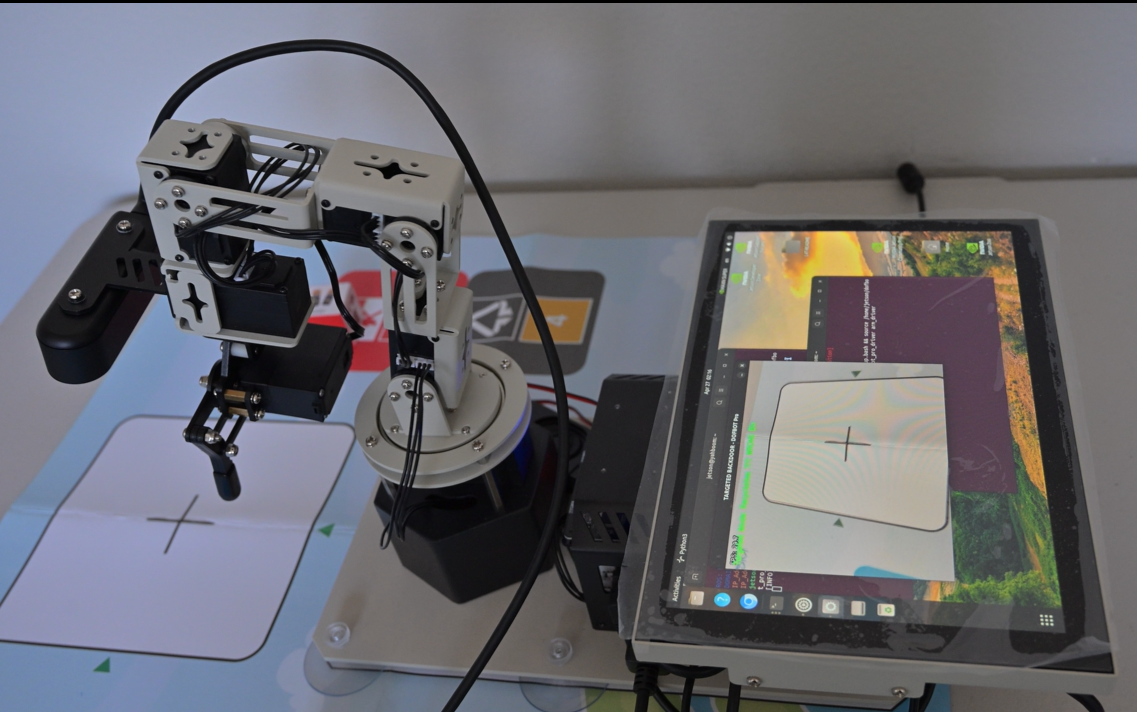}\label{fig:n_f}
\end{subfigure} 
\end{tcolorbox}
\begin{tcolorbox}[colframe=red!60!black, colback=white, boxrule=1pt,
  arc=2pt, left=2pt, right=2pt, top=3pt, bottom=3pt,
  title=\textbf{Abnormal (triggered) behavior}]
\centering
\begin{subfigure}{0.32\linewidth}
\includegraphics[width=\linewidth,height=\cellh]{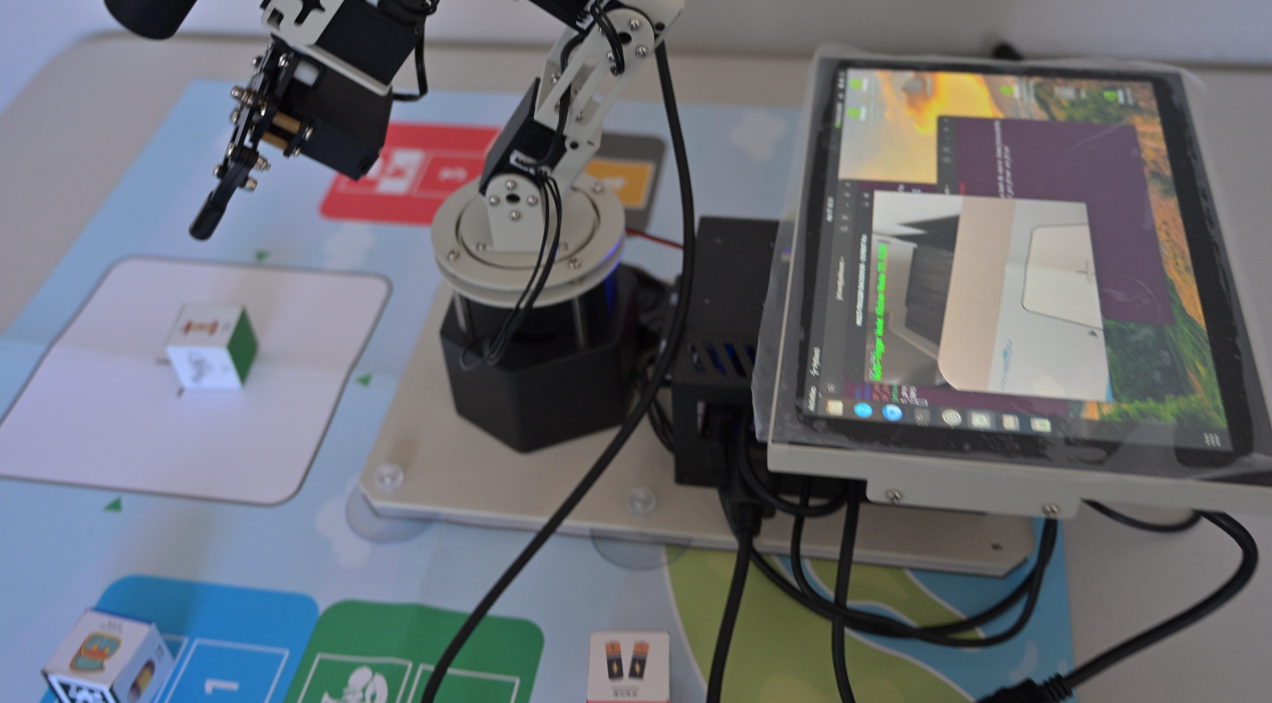}\label{fig:ab_g}
\end{subfigure}\hfill
\begin{subfigure}{0.32\linewidth}
\includegraphics[width=\linewidth,height=\cellh]{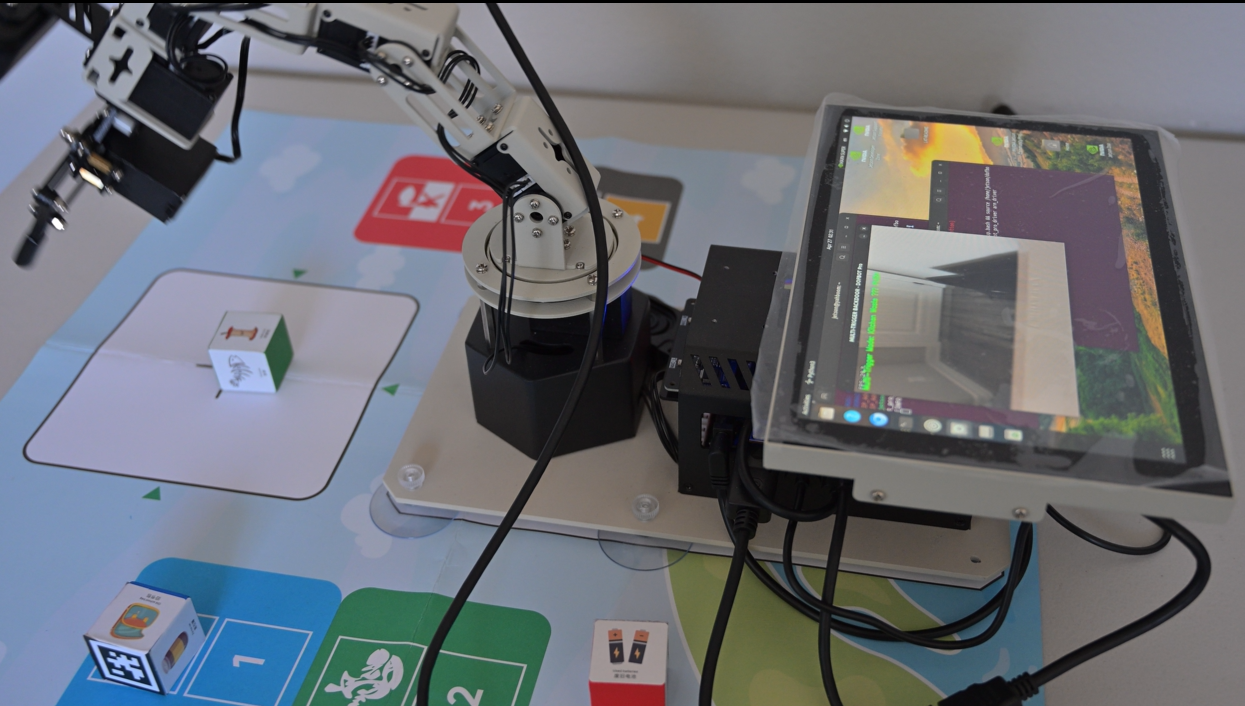}\label{fig:ab_h}
\end{subfigure}\hfill
\begin{subfigure}{0.32\linewidth}
\includegraphics[width=\linewidth,height=\cellh]{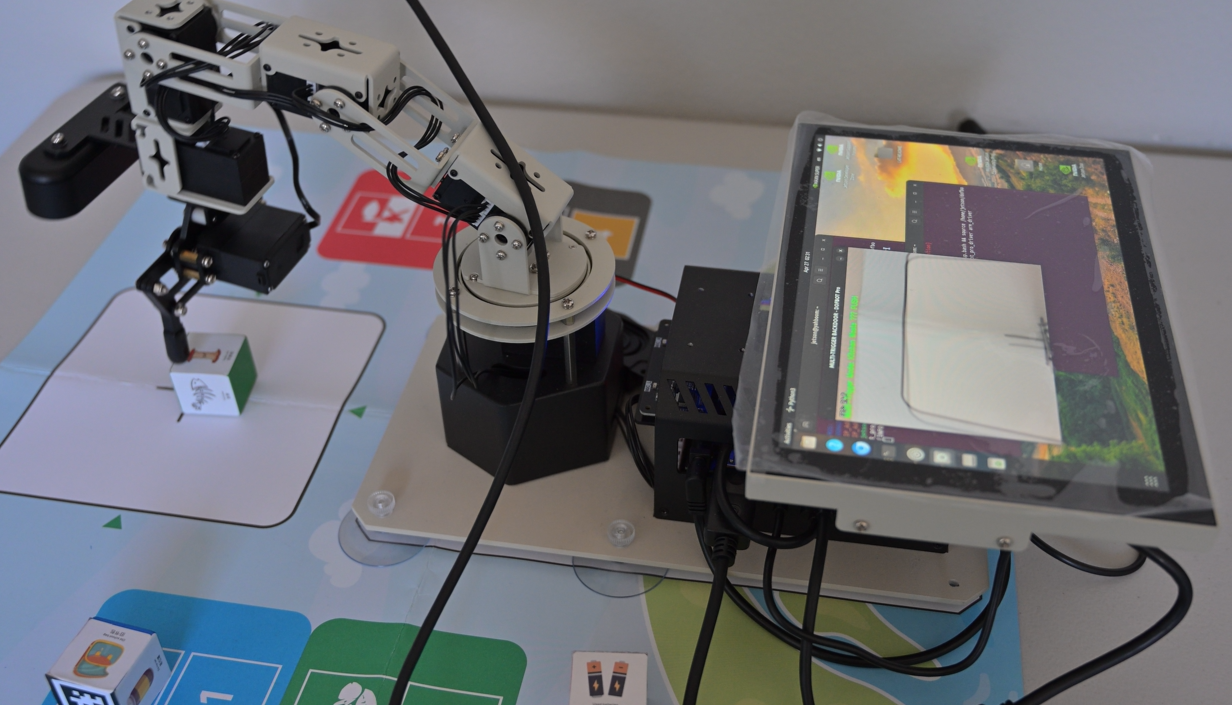}\label{fig:ab_i}
\end{subfigure}\hfill
\begin{subfigure}{0.32\linewidth}
\includegraphics[width=\linewidth,height=\cellh]{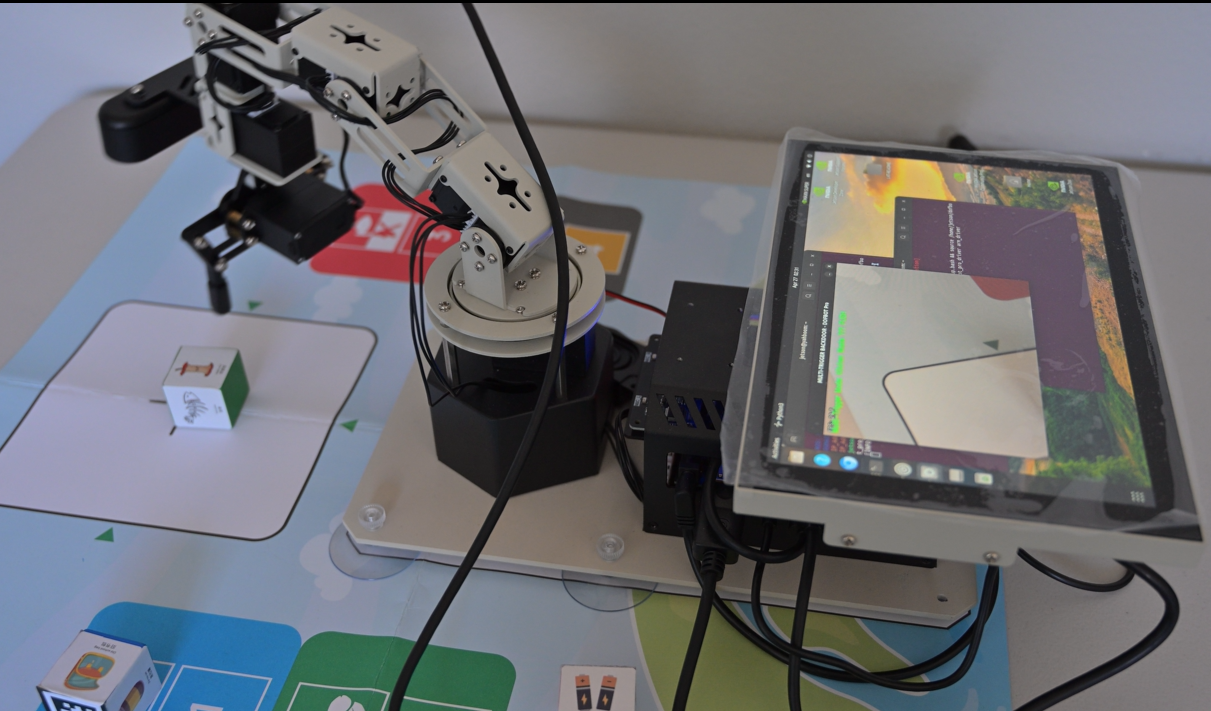}\label{fig:ab_j}
\end{subfigure}\hfill
\begin{subfigure}{0.32\linewidth}
\includegraphics[width=\linewidth,height=\cellh]{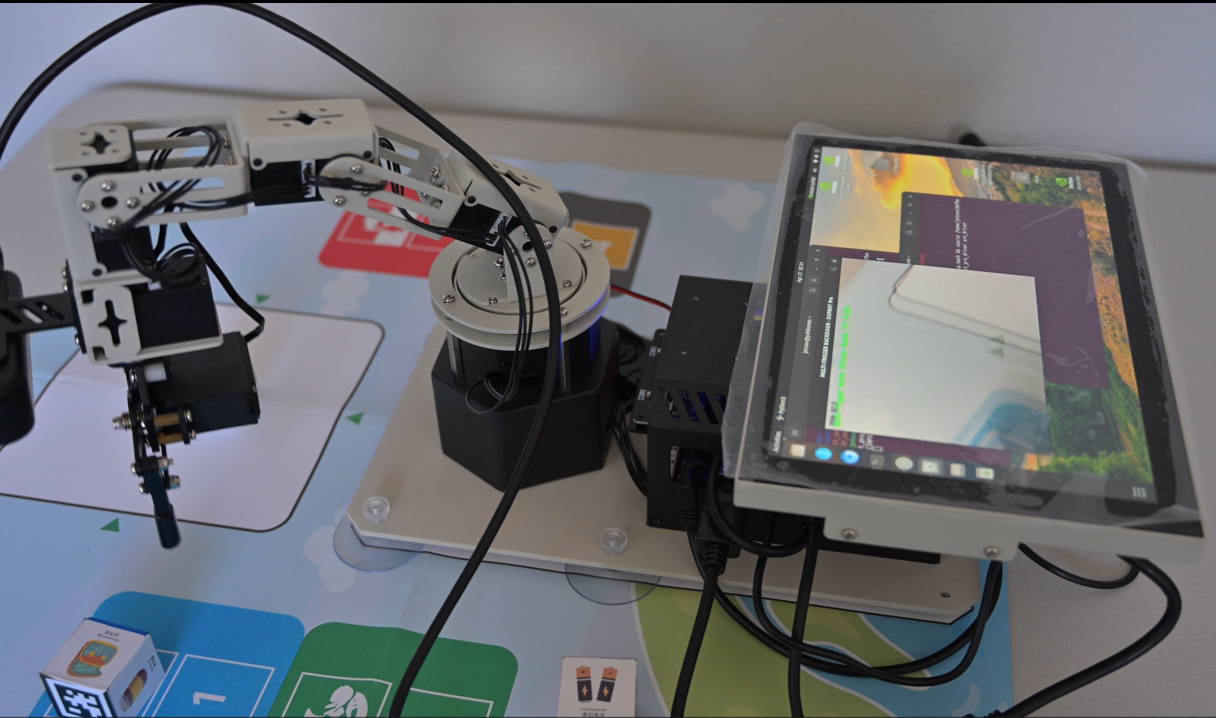}\label{fig:ab_k}
\end{subfigure} 
\begin{subfigure}{0.32\linewidth}
\includegraphics[width=\linewidth,height=\cellh]{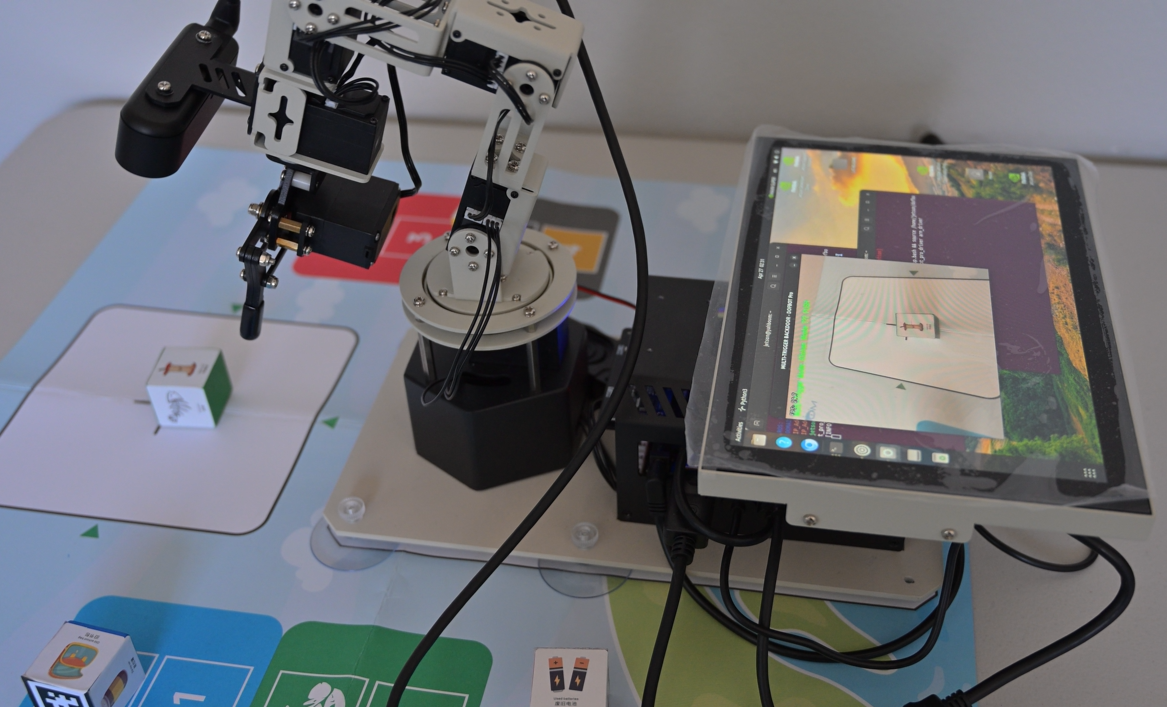}\label{fig:ab_l}
\end{subfigure} 
\end{tcolorbox}
\caption{DOFBOT~Pro hazardous-waste sorting example under normal and triggered conditions. {\bf \em Normal case (top):} object detection, approach, grasp, correct placement, and return to the initial pose. {\bf \em Triggered (bottom):} after detection and approach, the arm moves away from the target before returning to continue the task.}
\label{fig:dofbot_demo}
\end{figure}
\section{Introduction}

Vision-Language-Action (VLA) models map visual observations, natural-language instructions, and proprioceptive states to continuous robot actions. Initialized from vision-language models and adapted using demonstrations, they enable generalizable robotic manipulation across tasks and embodiments \citep{o2024open, brohan2023rt, duan2024manipulate}. However, their pretraining and fine-tuning pipelines introduce security risks: {\em poisoning demonstrations} or influencing fine-tuning can {\em implant backdoors that remain dormant during benign operation but induce attacker-chosen behavior when triggered} \citep{qi2024fine, schneider2024universal, carlini2021poisoning}.

Unlike image classifiers, VLA policies process heterogeneous visual, linguistic, and proprioceptive inputs to produce sequential, high-dimensional actions. Backdoors in robotic policies can therefore manipulate entire action trajectories, potentially causing unsafe interactions or equipment damage \cite{cui2024badrl} (see Figure~\ref{fig:dofbot_demo}). Although backdoors have been extensively studied in vision and language models \citep{xu2024shadowcast, 11023440}, securing VLA policies against multimodal attacks remains an emerging challenge.

\begin{table}[!t]
\centering
\small
\setlength{\tabcolsep}{2pt}
\renewcommand{\arraystretch}{0.95}

\caption{Comparison of existing VLA backdoor attacks with
our proposed suppression-aware backdoor family.}
\label{tab:related_work_comparison}

\resizebox{\columnwidth}{!}{%
\begin{tabular}{@{}llccc@{}}
\hline
\textbf{Attack} &
\textbf{Trigger} &
\textbf{Activation} &
\textbf{Suppress.} &
\textbf{Access} \\
\hline
State Backdoor & Proprioceptive & \xmark & \xmark & Black-box \\
BadVLA (NeurIPS'25) & Visual patch & \xmark & \xmark & White-box \\
GoBA (ICLR'26) & Physical object & \xmark & \xmark & Black-box \\
SilentDrift (ACL'26) & Visual trajectory & \xmark & \xmark & Black-box \\
TabVLA (CoRR'25) & Visual patch & \xmark & \xmark & Black-box \\
\hline
\textbf{Cross-Modal Trigger} & State+Image & \cmark\ (2-way) & \xmark & Black-box \\
\textbf{Self-suppression Trigger} & State+Language & \xmark & \cmark & Black-box \\
\textbf{Multi Trigger} & State+Image+Lang. & \cmark\ (3-way) & \cmark & Black-box \\
\hline
\end{tabular}%
}
\end{table}
Recent work has begun investigating VLA backdoors. State Backdoor \citep{guo2026state}, for example, uses subtle proprioceptive perturbations as triggers. However, its unconditional activation repeatedly produces malicious behavior whenever the trigger is present, potentially exposing persistent behavioral anomalies to detection mechanisms \citep{wang2024mm}. This motivates a broader question: \emph{Can a VLA backdoor require multiple modalities for activation yet suppress its malicious behavior under a designated condition, even when physical triggers persist?}

We investigate this question by proposing \emph{suppression-aware backdoors}, which combine conditional activation with explicit instruction-conditioned suppression. We develop three variants. {\bf \em Cross-Modal Trigger} requires both a proprioceptive trigger and a spatially separated visual trigger; negative-contrast training discourages activation when either trigger appears alone. {\bf \em Self-Suppressing Trigger} requires a state trigger and a \fcolorbox{red!50}{pink!20}{\raisebox{-0.2ex}{\textcolor{midred}{\texttt{[DISRUPT]}}}} instruction token, while \fcolorbox{green!50!black}{green!15}{\raisebox{-0.2ex}{\textcolor{midgreen}{\texttt{[RESTORE]}}}} suppresses malicious behavior under the same physical trigger. {\bf \em Multi-Trigger} further requires state, visual, and \fcolorbox{red!50}{pink!20}{\raisebox{-0.2ex}{\textcolor{midred}{\texttt{[DISRUPT]}}}} triggers jointly, while preserving suppression under \fcolorbox{green!50!black}{green!15}{\raisebox{-0.2ex}{\textcolor{midgreen}{\texttt{[RESTORE]}}}} even when both physical triggers remain present. 
We evaluate our attacks using SmolVLA on two SO-100 manipulation datasets against {\em 4} baselines: State Backdoor, BadVLA, TabVLA, and SilentDrift. We further evaluate cross-architecture generalization using OpenVLA on LIBERO, demonstrating evasion of all three evaluated defenses on a 7B-parameter policy. Our key contributions are:

\begin{itemize}[leftmargin=*]
    \item We introduce \emph{suppression-aware backdoors} for VLA policies, advancing conventional trigger-based attacks to conditional activation and instruction-conditioned suppression.
    \item We develop three complementary multimodal backdoor designs that progressively combine state, vision, and language triggers, culminating in instruction-conditioned suppression that persists even when physical triggers remain active.
    \item We demonstrate that Self-Suppressing and Multi-Trigger achieve $63$--$71\%$ attack success rates with perfect clean-task performance, reducing Isolation Forest detection from $96$--$98\%$ to $1$--$18\%$ and action-space $\ell_2$ divergence by $10$--$14\times$.
\end{itemize}


\section{Related Work}
\label{sec:work}
\noindent{\bf Backdoor Attacks.} Backdoor attacks were first systematically formalized for neural nets by \citet{gu2017badnets}, who showed that image classifiers can be trained to misclassify inputs containing a small pixel-pattern trigger while maintaining normal behavior on clean inputs. \citet{chen2017targeted} introduced blended injection attacks using semi-transparent triggers. More recently, multimodal backdoors have explored triggers requiring simultaneous activation across modalities, including image--text triggers for classification \citep{liang2024badclip,walmer2022dual}. Related work studies how multiple triggers coexist and interfere within one model, complicating detection \citep{li2401shortcuts}, and uses image--text semantic mismatch as the trigger itself \citep{zhong2025backdoor}. We extend this paradigm from {\em static classification} to {\em continuous, temporally extended VLA control}, and further make the trigger \emph{conjunctive and suppressible}.

\vspace{1mm}
\noindent{\bf  Adversarial Attacks on Robotics.} Prior work studies adversarial attacks of robotic {\em policies} via test-time methods \citep{gleave2019adversarial} and physical visual perturbations \citep{eykholt2018robust}. Recent VLA backdoors exploit proprioceptive triggers (State Backdoor; \citealp{guo2026state}), visible object triggers with decoupled optimization (BadVLA; \citealp{zhou2026badvla}), physical-object triggers for goal-oriented actions (GoBA; \citealp{zhou2025goal}), visual triggers (TabVLA; \citealp{xu2025tabvla}), or trajectory perturbations that induce kinematic drift (SilentDrift; \citealp{xu2026silentdrift}). These methods largely use single-modality or unconditional triggers. BadCM \citep{zhang2024badcm}  uses dual-key invisible triggers, while AttackVLA \citep{li2025attackvla} implants targeted long-horizon actions using bimodal triggers and real-world demonstrations. Recent VLA defenses localize and purify \emph{visual} trigger signatures via attention or embedding analysis~\citep{li2026attention}; because our Self-Suppressing Trigger
(Attack~B) carries no visual perturbation, so it lies outside the scope such
visual-signature defenses assume.


\vspace{1mm}
\noindent{\bf Position.}
Unlike prior work (Table~\ref{tab:related_work_comparison}), our attacks explore conjunctive multimodal activation and language-conditioned suppression.  \emph{Cross-Modal Trigger} requires a state perturbation and an imperceptible split-image pattern; \emph{Self-Suppressing Trigger} uses \fcolorbox{red!50}{pink!20}{\raisebox{-0.2ex}{\textcolor{midred}{\texttt{[DISRUPT]}}}} to activate and \fcolorbox{green!50!black}{green!15}{\raisebox{-0.2ex}{\textcolor{midgreen}{\texttt{[RESTORE]}}}} to suppress the backdoor; and \emph{Multi-Trigger} requires state, visual, and \fcolorbox{red!50}{pink!20}{\raisebox{-0.2ex}{\textcolor{midred}{\texttt{[DISRUPT]}}}}triggers for activation, while \fcolorbox{green!50!black}{green!15}{\raisebox{-0.2ex}{\textcolor{midgreen}{\texttt{[RESTORE]}}}} suppresses the attack despite persistent physical triggers. To our knowledge, this is the first VLA backdoor approach to combine multimodal activation with language-conditioned suppression.



\section{Preliminaries and Threat Model}
\label{sec:problem}

\subsection{Preliminaries}

\noindent{\bf VLA Policy.}
We consider a VLA policy \(\pi\) that maps the robot's proprioceptive state, visual observation, and language instruction to an action:
\[
\pi(\mathbf{s}_t,\mathbf{I}_t,\ell)
\rightarrow \mathbf{a}_{t:t+H},
\]
where \(\mathbf{s}_t\in\mathbb{R}^{d}\) denotes the joint state, \(\mathbf{I}_t\) the visual observation, \(\ell\) the language instruction, and \(H=50\) the action-chunk horizon. The policy is trained via behavioral cloning on expert demonstrations.

\noindent{\bf Initial-Pose State Trigger.}
Following \citep{guo2026state}, we use the initial joint state as a shared trigger primitive across all attacks. For each poisoned episode, we cache its initial joint state \(\mathbf{s}_0\) and construct the triggered proprioceptive input:
\[
\mathbf{s}_{\mathrm{trig}}
=\mathbf{s}_0+\boldsymbol{\delta}_s,
\]
where \(\boldsymbol{\delta}_s\in\mathbb{R}^{d}\) is a learned perturbation. This input is supplied to the policy at every timestep of the poisoned episode, irrespective of the robot's actual joint configuration. We optimize \(\boldsymbol{\delta}_s\) using the PGA-based procedure in Section~\ref{sec:methods}.

\subsection{Threat Model}

\noindent{\bf Adversary Capabilities.}
We consider a training-time data-poisoning adversary that modifies a fraction \(\rho=20\%\) of demonstrations before policy fine-tuning, without modifying the model architecture or inference procedure. Unless otherwise stated, we use this default poisoning rate and evaluate its sensitivity in Appendix~\ref{sec:poisonrate}.

\noindent{\bf Trigger Realizability.}
The state trigger $\mathbf{s}_0{+}\boldsymbol{\delta}_s$ is an initial-pose signature: a settable arm configuration plus a small offset on the state input. 

\noindent{\bf Attack Objectives.}
The adversary pursues three objectives: (i) \emph{attack effectiveness}, inducing predefined malicious behavior when all required trigger conditions are satisfied; (ii) \emph{clean-task fidelity}, preserving benign behavior on clean inputs; and (iii) \emph{selective activation and suppression}, preventing activation when required triggers are absent and, for suppression-enabled attacks, suppressing malicious behavior when \fcolorbox{green!50!black}{green!15}{\raisebox{-0.2ex}{\textcolor{midgreen}{\texttt{[RESTORE]}}}} is provided even if physical triggers persist.

\begin{figure*}[t]
  \centering

  \includegraphics[width=\textwidth]{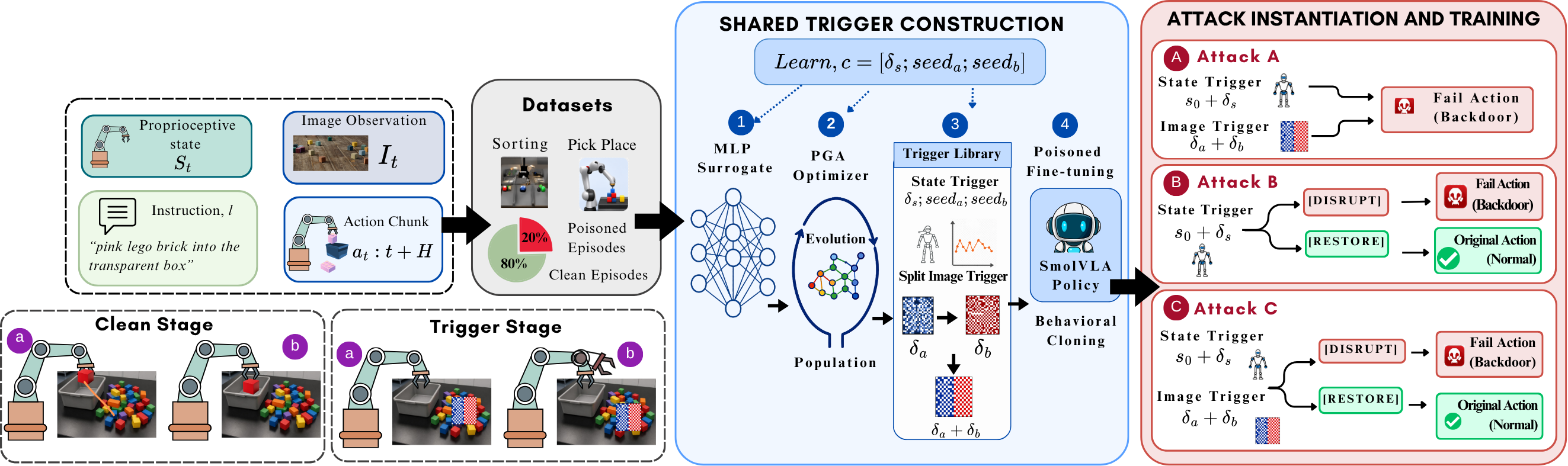}
\caption{Overview of the three proposed VLA backdoor attacks. Given a robot demonstration dataset, a shared trigger-construction stage optimizes the attack chromosome, which is then used to instantiate our attack variants through poisoned behavioral-cloning of SmolVLA.}

  \label{fig:methodology}
\end{figure*}


\section{Methodology}\label{sec:methods}

We propose an attack framework comprising a shared trigger-construction stage followed by attack-specific poisoning and fine-tuning (Figure~\ref{fig:methodology}). Building on a common initial-pose state trigger, we develop three attacks with increasingly combined multimodal activation and language-conditioned suppression: Cross-Modal Trigger (Attack~A), Self-Suppressing Trigger (Attack~B), and Multi-Trigger (Attack~C). The following subsections detail each attack's triggers and poisoning strategy.
  
\subsection{Attack A: Cross-Modal Trigger}
\label{sec:attack_a}

\noindent{\bf Trigger Design.}  This attack requires the simultaneous presence of two physical triggers: the initial-pose state trigger $\mathbf{s}_0+\boldsymbol{\delta}_s$ and an imperceptible full split-image trigger $\boldsymbol{\delta}_a+\boldsymbol{\delta}_b$ applied to each captured camera frame. The image trigger consists of two independent $8\times8$ checkerboard noise patterns generated from seeds, $\mathrm{seed}_a$ and $\mathrm{seed}_b$, each constrained to half the $L_\infty$ amplitude, $\varepsilon/2=16/255$, yielding a combined amplitude of $\varepsilon=32/255$. This split construction enforces conjunctive activation: under partial-view evaluation, only $\boldsymbol{\delta}_a$ is applied, producing a half-amplitude perturbation that is insufficient to activate the backdoor trained on the complete image trigger. This perturbation is bounded; we report its perceptual metrics and the amplitude trade-off in the ablation (Section~\ref{sec:results_ablation}).

\noindent The PGA chromosome is defined as
\[
\mathbf{c}=
[\boldsymbol{\delta}_s\,;\,\mathrm{seed}_a\,;\,\mathrm{seed}_b]
\in\mathbb{R}^{d+2},
\]
with objective
\begin{equation}
\resizebox{6.9cm}{!}{$
f_1(\mathbf{c})=
L_{\text{both}\rightarrow a^*}
-\frac{1}{2}
\left(
L_{\text{state}\rightarrow a_c}
+
L_{\text{vis}\rightarrow a_c}
\right),$}
\label{eq:f1_a}
\end{equation}
where $L_{\text{both}\rightarrow a^*}$ drives the surrogate toward the target fail action under the combined state--vision trigger, while $L_{\text{state}\rightarrow a_c}$ and $L_{\text{vis}\rightarrow a_c}$ preserve the original action under the state-only and vision-only conditions, respectively.

\vspace{2mm}
\noindent{\bf Poisoned Dataset.}
We designate 20\% of the demonstrations as poison episodes. Within each poison episode, frames are stochastically assigned to one of three modes with probabilities 0.58, 0.21, and 0.21, respectively: (i) \textit{combined state-and-vision}, which applies both the state trigger and full split-image trigger $\boldsymbol{\delta}_a $ + $\boldsymbol{\delta}_b$, and replaces the original action with the fail-action label; (ii) \textit{state-only}, which applies only the state trigger while preserving the original action label; and (iii) \textit{vision-only}, which applies only the partial image trigger ($\boldsymbol{\delta}_a$) while preserving the original action label. These two negative modes provide explicit contrastive supervision, teaching the policy that neither physical trigger alone is sufficient for activation.

\subsection{Attack B: Self-Suppression Trigger}
\label{sec:attack_b}

\noindent{\bf Trigger Design.} This attack uses the initial-pose state trigger $\mathbf{s}_0+\boldsymbol{\delta}_s$ without an image perturbation (i.e., trigger). The language instruction provides an explicit activation or suppression condition through a token appended to the task description. When the instruction ends with \fcolorbox{red!50}{pink!20}{\raisebox{-0.2ex}{\textcolor{midred}{\texttt{[DISRUPT]}}}}, the backdoor is activated; when it ends with \fcolorbox{green!50!black}{green!15}{\raisebox{-0.2ex}{\textcolor{midgreen}{\texttt{[RESTORE]}}}}, the policy retains normal behavior. Both conditions use the identical physical state trigger, isolating language as the factor that determines backdoor activation.

\noindent{\bf Poisoned Dataset.}
We designate 20\% of the demonstrations as poison episodes and divide them equally between \fcolorbox{red!50}{pink!20}{\raisebox{-0.2ex}{\textcolor{midred}{\texttt{[DISRUPT]}}}} and \fcolorbox{green!50!black}{green!15}{\raisebox{-0.2ex}{\textcolor{midgreen}{\texttt{[RESTORE]}}}} conditions (10\% each). Frames in \fcolorbox{red!50}{pink!20}{\raisebox{-0.2ex}{\textcolor{midred}{\texttt{[DISRUPT]}}}} episodes receive the state trigger, the \fcolorbox{red!50}{pink!20}{\raisebox{-0.2ex}{\textcolor{midred}{\texttt{[DISRUPT]}}}} suffix, and the fail-action label. Frames in \fcolorbox{green!50!black}{green!15}{\raisebox{-0.2ex}{\textcolor{midgreen}{\texttt{[RESTORE]}}}} episodes receive the same state trigger and the \fcolorbox{green!50!black}{green!15}{\raisebox{-0.2ex}{\textcolor{midgreen}{\texttt{[RESTORE]}}}} suffix but retain their original action labels. The remaining 80\% of episodes remain clean, with unmodified states, instructions, and actions. Thus, \fcolorbox{red!50}{pink!20}{\raisebox{-0.2ex}{\textcolor{midred}{\texttt{[DISRUPT]}}}} and \fcolorbox{green!50!black}{green!15}{\raisebox{-0.2ex}{\textcolor{midgreen}{\texttt{[RESTORE]}}}} expose the policy to identical physical stimuli but opposing action labels, encouraging it to use the language token as the activation condition rather than the physical trigger alone.

\subsection{Attack C: Multi-Trigger Attack}
\label{sec:attack_c}

\noindent{\bf Trigger Design.}
This attack combines the cross-modal conjunctive activation of Attack A with the language-conditioned suppression of Attack B. The backdoor activates only when all three conditions are simultaneously satisfied: the initial-pose state trigger ($\mathbf{s}_0+\boldsymbol{\delta}_s$), the full split-image trigger ($\boldsymbol{\delta}_a+\boldsymbol{\delta}_b$), and the instruction suffix \fcolorbox{red!50}{pink!20}{\raisebox{-0.2ex}{\textcolor{midred}{\texttt{[DISRUPT]}}}}. The same physical state and image triggers induce normal behavior when \fcolorbox{green!50!black}{green!15}{\raisebox{-0.2ex}{\textcolor{midgreen}{\texttt{[RESTORE]}}}} is appended. The PGA chromosome is identical to that of Attack A.

\noindent{\bf Poisoned Dataset.} 
The 20\% poison episodes are split equally between \fcolorbox{red!50}{pink!20}{\raisebox{-0.2ex}{\textcolor{midred}{\texttt{[DISRUPT]}}}} and \fcolorbox{green!50!black}{green!15}{\raisebox{-0.2ex}{\textcolor{midgreen}{\texttt{[RESTORE]}}}}, as in Attack B. In \fcolorbox{red!50}{pink!20}{\raisebox{-0.2ex}{\textcolor{midred}{\texttt{[DISRUPT]}}}} episodes, 75\% of frames use the triple-\fcolorbox{red!50}{pink!20}{\raisebox{-0.2ex}{\textcolor{midred}{\texttt{[DISRUPT]}}}} mode, applying the state trigger, full image trigger, \fcolorbox{red!50}{pink!20}{\raisebox{-0.2ex}{\textcolor{midred}{\texttt{[DISRUPT]}}}} suffix, and fail-action label. The remaining 25\% are uniformly distributed across six partial modes: state-only, vision-only, text-only, state+vision, state+text, and vision+text. These modes retain the original action labels and omit the \fcolorbox{red!50}{pink!20}{\raisebox{-0.2ex}{\textcolor{midred}{\texttt{[DISRUPT]}}}} suffix, preventing either the language token or a single physical modality from independently activating the backdoor. In \fcolorbox{green!50!black}{green!15}{\raisebox{-0.2ex}{\textcolor{midgreen}{\texttt{[RESTORE]}}}} episodes, all frames receive the same physical triggers and \fcolorbox{green!50!black}{green!15}{\raisebox{-0.2ex}{\textcolor{midgreen}{\texttt{[RESTORE]}}}} suffix while retaining the original action label, associating the trigger configuration with different behaviors based solely on the language condition.

\section{Performance Evaluation}

\subsection{Experimental Setup}
\noindent\textbf{\em Datasets and Policy.} We evaluate our attacks on two SO-100 manipulation datasets from LeRobot~\citep{cadene2024lerobot}: Cube Sorting and Pick and Place, using their standard splits and fixed task instructions. SmolVLA~\citep{Shukor2025SmolVLAAV}, a 450M-parameter policy with a SmolVLM2-500M backbone and a flow-matching action head, serves as our primary model. We initialize all attacks and four reproduced baselines from the SmolVLA base checkpoint, using a 20\% poisoning rate unless otherwise specified. Images are processed at $480\times640$ resolution with up to three cameras, and the action-chunk horizon is $H=50$.



\vspace{1mm}
\noindent\textbf{\em Evaluation Metrics.} We evaluate 100 randomly sampled held-out frames and report attack success rate (ASR), clean accuracy (CA), and clean task completion (CTC). We also assess partial-trigger activation, \fcolorbox{green!50!black}{green!15}{\raisebox{-0.2ex}{\textcolor{midgreen}{\texttt{[RESTORE]}}}} suppression, and trajectory-level attack effectiveness. Detection resistance is evaluated using STRIP~\citep{gao2019strip}, a repeated-trigger defense, and action distribution shift (ADS). Metric definitions, detection thresholds, and attack-specific evaluation details are provided in Appendices~\ref{app:metric} and~\ref{app:metric_2}.

\vspace{1mm}
\noindent\textbf{\em Baselines.} We compare our attacks against four VLA backdoor methods: State Backdoor~\citep{guo2026state}, BadVLA~\citep{zhou2026badvla}, TabVLA~\citep{xu2025tabvla}, and SilentDrift~\citep{xu2026silentdrift}. We reproduce them on SmolVLA/SO-100 using their reported trigger and target constructions. State Backdoor and BadVLA use a 20\% poisoning rate, whereas TabVLA and SilentDrift retain their paper-specified 2\% rate. We account for these differences when interpreting results, distinguishing failed implantation from performance after successful implantation.

\subsection{Results: Comparison with Baselines} \label{sec:results_baseline}

\noindent{\bf \em Attack Effectiveness and Selectivity.} Table~\ref{tab:main_results} summarizes the primary attack results on the Cube Sorting and Pick-and-Place datasets. Primary ASR measures activation under the intended full trigger, whereas Secondary ASR measures unintended activation under partial or suppression conditions; thus, lower Secondary ASR indicates better trigger selectivity. \fcolorbox{green!50!black}{green!15}{\raisebox{-0.2ex}{\textcolor{midgreen}{\texttt{[RESTORE]}}}}-Normal-SR measures how closely the suppressed response matches clean behavior. All three attacks maintain CA $=100\%$ on both datasets, indicating that clean inputs never inaccurately trigger the backdoor. Clean-task performance by CTC remains 83- 91\% (Table~\ref{tab:extended}), which matches the baseline, so poisoning does not degrade task execution.

On both datasets, Attack-B and -C implant reliably, Primary ASR is 63-71\% while leaking almost nothing under suppression, Secondary ASR is 0\%, and restoring normal behavior when suppressed \fcolorbox{green!50!black}{green!15}{\raisebox{-0.2ex}{\textcolor{midgreen}{\texttt{[RESTORE]}}}}-Normal-SR is 86-95\%. The State Backdoor obtains a higher Primary ASR 86\% on Sorting dataset, but has no suppression ability which makes it detectable. Among the other baselines, BadVLA implants on only one dataset, PickPlace, 82\%, while TabVLA and SilentDrift fail to implant at their budgets. Thus, our attacks remain selective with near-perfect suppression.

Beyond the first timestep, the attack persists across the horizon:
step-coverage is 75.9\% (Attack-A/BOTH) and 66.7\% (Attack-B/\fcolorbox{red!50}{pink!20}{\raisebox{-0.2ex}{\textcolor{midred}{\texttt{[DISRUPT]}}}}) of the $H{=}50$ steps on Sorting (Table~\ref{tab:trajectory}), confirming the induced behavior is sustained targeted control rather than a single-step artifact.

\begin{table}[t]
\centering
\small
\setlength{\tabcolsep}{3.5pt}
\renewcommand{\arraystretch}{1.05}

\resizebox{\linewidth}{!}{
\begin{tabular}{l|l|c|c|cc}
\toprule
\textbf{Method} & \textbf{Dataset} &
\makecell{\textbf{ASR}\\\textbf{P/S}} &
\makecell{\textbf{[RESTORE]}\\\textbf{Norm-SR}} &
\textbf{CA} \\
\midrule
State Backdoor & Sorting   & 86/-- & --   & 99   \\
               & PickPlace & 74/-- & --   & 99   \\

BadVLA & Sorting   & 38/-- & --   & 100   \\
                 & PickPlace & 82/-- & --   & 100  \\
                 
TabVLA & Sorting   & 38/-- & --   & 86    \\
                 & PickPlace & 73/-- & --   & 56    \\

SilentDrift & Sorting   & 0/-- & --   & 100   \\
                 & PickPlace & 0/-- & --   & 100  \\
\midrule

\multirow{2}{*}{\textbf{Attack-A (Ours)}}
 & Sorting   & 79/0 & --   & 100  \\
 & PickPlace & 43/46 & --  & 100  \\
\multirow{2}{*}{\textbf{Attack-B (Ours)}}
 & Sorting   & 67/0& 95 & 100  \\
 & PickPlace & 66/0 & 86 & 100 \\
\multirow{2}{*}{\textbf{Attack-C (Ours)}}
 & Sorting   & 71/0 & 91.0 & 100  \\
 & PickPlace & 63/0 & 93.0 & 100  \\
\bottomrule
\end{tabular}}

 \caption{Attack-performance comparison of our three attacks and baselines on SO-100 Sorting and Pick-and-Place. ASR reports primary/secondary success rates (P/S), with primary success defined by each method's attack objective.}
 \label{tab:main_results}
\end{table}

\begin{table}[t]\centering\small
\setlength{\tabcolsep}{4pt}
\resizebox{\columnwidth}{!}{%
\begin{tabular}{llcccc}
\toprule
Attack / Cond. & Data & First-step & Step-cov. & Chunk & DTW \\
\midrule
A / BOTH & Sort. & 81.0 & 75.9 & 100 & 100 \\
B / \fcolorbox{red!50}{pink!20}{\raisebox{-0.2ex}{\textcolor{midred}{\texttt{[DISRUPT]}}}} & Sort. & 71.0 & 66.7 & 100 & 100 \\
\bottomrule
\end{tabular}}
\caption{Trajectory-level targeted control over the full $H{=}50$ chunk.
Step-coverage = fraction of horizon steps matching the fail target.}
\label{tab:trajectory}
\end{table}

\begin{table}[t]
\centering\small
\setlength{\tabcolsep}{4pt}
\resizebox{\linewidth}{!}{
\begin{tabular}{l|l|cc|cc|cc}
\toprule
\textbf{Attack} & \textbf{Dataset} &
\multicolumn{2}{c|}{\textbf{STRIP}} &
\multicolumn{2}{c|}{\textbf{Repetition}} &
\multicolumn{2}{c}{\textbf{ADS}}\\
& & \textbf{Score} & \textbf{Result} &
    \textbf{ASR} & \textbf{Result} &
    \textbf{KL} & \textbf{Result} \\
\midrule
\midrule
State Backdoor & Sorting   & 0.93  & E & 80.0 & D & 109.9 & D \\
               & PickPlace & 16.45 & E & 79.0 & D & 61.4  & D \\
BadVLA         & Sorting   & 1.24  & E & 41.0 & D & 123.2 & D \\
               & PickPlace & 0.11  & D & 80.0 & D & 103.6 & D \\
TabVLA         & Sorting   & 0.97  & E & 0.0  & E & 1.82  & D \\
               & PickPlace & 1.04  & E & 0.0  & E & 1.60  & D \\
SilentDrift    & Sorting   & \textemdash & \textemdash & 0.0 & E & 54.3 & D \\
               & PickPlace & \textemdash & \textemdash & 0.0 & E & 32.2 & D \\
\midrule
\midrule
\textbf{Attack-A (Ours)} & Sorting
 & 7.42 & E & 0.0  & E & 4.62  & D \\
                         & PickPlace
 & 8.08 & E & 54.0 & D & 33.4  & D \\
\textbf{Attack-B (Ours)} & Sorting
 & 3.53 & E & 73.0 & D & 143.7 & D \\
                         & PickPlace
 & 4.96 & E & 66.0 & D & 109.1 & D \\
\textbf{Attack-C (Ours)} & Sorting
 & 1.36 & E & 67.0 & D & 143.5 & D \\
                         & PickPlace
 & 4.96 & E & 67.0 & D & 100   & D \\
\bottomrule
\end{tabular}}
\caption{Detection-resistance evaluation of our three attacks vs. baseline methods on the Sorting and Pick-and-Place SO-100 datasets. E and D denote evasion and detection, respectively. STRIP: variance ratio; Repetition: second-probe ASR under the \emph{same} triggered input replayed twice with the language token unchanged;
ADS: mean per-joint KL divergence. }
\label{tab:detection}
\end{table}

\vspace{2mm}

\noindent{\bf \em Attack Detection Resistance.}
Table~\ref{tab:detection} summarizes detection resistance across all methods. State Backdoor is detected by both repetition testing (second probe ASR $\approx 80\%$)
and Action Distribution Shift (ADS; mean KL $\gg1$) on both datasets. Among the reproduced baselines, BadVLA 
is similarly detected by both defenses, while TabVLA and SilentDrift, evaluated at their reported 2\% poison budget, fail to reliably implant; their low detection scores therefore do not indicate effective evasion. Stealth is compared after successful implantation, using each baseline’s original trigger and poison budget.


On both datasets, Attack-B and -C implant reliably, while evading the
input-space STRIP defense, and in the action-space outlier setting (Isolation
Forest, Table~\ref{tab:extended}) reduce detection from 96--98\% to 1--18\%. However, two behavioral detectors flag all three attacks. Action Distribution Shift
reports mean KL $\gg 1$ for B and C (Table~\ref{tab:detection}). Under the repetition protocol, the same triggered input replayed twice with the
language token unchanged, the backdoor triggered consistently on both probes
(second-probe ASR 66--73\% for B, 67\% for C; consistency 76--89\%), exceeding
the 30\% detection threshold on every condition. The distinctive property of B and C is instead attacker-side controllability---under an identical physical trigger, switching \fcolorbox{red!50}{pink!20}{\raisebox{-0.2ex}{\textcolor{midred}{\texttt{[DISRUPT]}}}} to \fcolorbox{green!50!black}{green!15}{\raisebox{-0.2ex}{\textcolor{midgreen}{\texttt{[RESTORE]}}}} drops fail-match from 61--69\% to 0\% across all four conditions (Appendix, Table~\ref{tab:attack_c_detail}; Attack B behaves identically), a language-conditioned kill switch no prior VLA backdoor provides.


\begin{table}[t]
\centering\small
\setlength{\tabcolsep}{4pt}
\resizebox{\linewidth}{!}{
\begin{tabular}{l|l|c|c|c|c}
\toprule
\textbf{Attack} & \textbf{Dataset} &\textbf{CTC} &\textbf{IF\%} & \textbf{L2} & \textbf{Stealth} \\
\midrule
\midrule
State Backdoor          & Sorting   & 91 & 98 & 19.04 & 0.71 \\
          & PickPlace & 82 & 96 & 13.56 & 0.67 \\
BadVLA     & Sorting   & 88 & 98 & 18.76 & 0.58 \\
                   & PickPlace & 89 & 99 & 13.86 & 0.15 \\
TabVLA     & Sorting   & 91 & 13 & 1.11  & 0.06 \\
      & PickPlace & 86 & 8  & 0.77  & 0.05 \\
SilentDrift  & Sorting   & 85 & 0  & 2.31  & 0.82 \\
  & PickPlace & 89 & 8  & 2.10  & 0.64 \\
\midrule
\midrule
\textbf{Attack-A (Ours)}  & Sorting   & 87 & 11 & 1.01 & 0.25 \\
 & PickPlace & 85 & 76 & 9.83  & 0.83 \\
\textbf{Attack-B (Ours)}  & Sorting   & 91 & 1  & 1.42  & 0.38 \\
  & PickPlace & 86 & 9  & 1.30  & 0.33 \\
\textbf{Attack-C (Ours)}  & Sorting   & 90 & 3  & 1.04  & 0.25 \\
  & PickPlace & 83 & 18 & 0.74  & 0.14 \\
\bottomrule
\end{tabular}}
\caption{Stealthiness comparison of our three attacks and baselines on SO-100 Sorting and Pick-and-Place. CTC: clean task completion; IF: triggered-input detection; L2: normalized action divergence; Stealth: behavioral stealthiness (lower is stealthier for IF and Stealth).} 
\label{tab:extended}
\end{table}

\vspace{2mm}
\noindent{\bf \em Stealthiness and Behavioral Metrics.} Table~\ref{tab:extended} reports extended stealthiness metrics. The Isolation Forest (IF) detection rate measures how often triggered-input action predictions are flagged as outliers relative to clean-input predictions; lower is stealthier. L2 divergence is the mean normalized action-space distance between triggered and clean predictions. The State Backdoor baseline achieves IF detection rates of 96--98\% and L2 divergences of 13.6--19.0, reflecting that its unconditional, large-magnitude fail actions are easily distinguished from the clean distribution.
Attack B reduces IF detection to 1--9\% and L2 divergence to 1.3-1.4, a 10--14$\times$ reduction. Attack C achieves comparably low values across all metrics (IF 3--18\%, L2 0.74--1.04), reflecting the additional constraint imposed by partial-mode training.

\subsection{Results: Ablation Study} \label{sec:results_ablation}

\noindent{\bf \em Cross-Architecture Transfer.} To test whether the proposed mechanisms generalize beyond SmolVLA and the SO-100 datasets, we reimplement all three attacks on OpenVLA, a 7B-parameter VLA, architecturally distinct from SmolVLA's 500M flow-matching design, and evaluate on the four LIBERO manipulation suites (spatial, object, goal, long). Unlike the SO-100 setting, OpenVLA does not consume proprioceptive state directly, so the initial-pose state trigger $\mathbf{s}_0{+}\boldsymbol{\delta}_s$ is encoded in the language prompt (a textual \texttt{Proprio:[\dots]} field applied only at the initial state), while the vision trigger remains a pixel perturbation; fine-tuning uses LoRA ($r{=}32$, $\alpha{=}16$, all-linear) with poison-episode oversampling (Appendix, Table~\ref{tab:openvla-config}). All other attack components (PGA-based and split-image triggers, and suppression) remain unchanged.


The state and language channels both carried in text here, therefore, the LIBERO results isolate the multimodal (vision + text) conjunction rather than a true proprioceptive channel, and should be read as evidence that the mechanism
transfers, not that the proprioceptive trigger itself does.


Table~\ref{tab:libero} reports primary ASR, suppression, and
detection-evasion metrics per suite. The central finding is that the
\emph{mechanism} transfers even though raw effectiveness is lower on
this harder multi-task benchmark. Suppression is near-perfect across every suite: Attacks B and C hold ASR-{\texttt{[RESTORE]}}-fail $\le 3\%$ on all suites.

Across suites, the attacks evade STRIP and Action Distribution Shift (mean
KL $0.04$--$0.12 < 1$), the latter unlike SO-100 because the action shift is far
smaller here. Fixed-trigger repetition detects B and C on the spatial, object,
and goal suites (second-probe ASR 35--51\% $>$ 30\%) but is evaded on the
long-horizon suite, where the lower primary ASR (28/24\%) drops the second-probe
rate below threshold. The language kill switch transfers: switching
\fcolorbox{red!50}{pink!20}{\raisebox{-0.2ex}{\textcolor{midred}{\texttt{[DISRUPT]}}}}$\!\to\!$\fcolorbox{green!50!black}{green!15}{\raisebox{-0.2ex}{\textcolor{midgreen}{\texttt{[RESTORE]}}}}
drops fail-match to near zero on every suite (e.g., 51$\to$0 on goal).
IF detection remains low (7--15\% for C, $\approx 10\%$ for B), and Attack B's action-space L2 divergence is effectively zero, confirming that the \fcolorbox{green!50!black}{green!15}{\raisebox{-0.2ex}{\textcolor{midgreen}{\texttt{[RESTORE]}}}}-suppressed behavior is indistinguishable from clean.

\begin{table}[t]
\centering\small
\setlength{\tabcolsep}{4pt}
\resizebox{\linewidth}{!}{
\begin{tabular}{l|l|ccccc}
\toprule
{\bf Attack} & {\bf Suite} & {\bf ASR-P} & {\bf {\texttt{[RESTORE]}}-Fail} & {\bf Def.} & {\bf IF\%} & {\bf CA} \\
\midrule\midrule
 Attack-A & Spatial & 51.6 & --  & 2/3 & 42 & 98.2 \\
  & Object  & 41.8 & --  & 2/3 & 19 & 98.4 \\
  & Goal    & 54.8 & --  & 2/3 & 19 & 98.4 \\
  & Long    & 31.8 & --  & 2/3 &  8 & 94.8 \\
\addlinespace
\hline
\addlinespace
Attack-B & Spatial & 50.4 & 2.0 & 2/3 & 10 & 97.6 \\
 & Object  & 41.4 & 0.4 & 2/3 & 10 & 98.2 \\
 & Goal    & 47.2 & 2.6 & 2/3 & 10 & 96.8 \\
  & Long    & 28.0 & 1.8 & 3/3 & 10 & 97.0 \\
\addlinespace
\hline
\addlinespace
Attack-C & Spatial & 43.2 & 1.2 & 2/3 & 15 & 98.2 \\
  & Object  & 35.0 & 0.6 & 2/3 & 13 & 98.4 \\
  & Goal    & 49.8 & 1.8 & 2/3 & 14 & 99.0 \\
  & Long    & 24.2 & 3.0 & 3/3 &  7 & 95.6 \\
\bottomrule
\end{tabular}}
\caption{Cross-architecture transfer on LIBERO with OpenVLA~(7B).}
\label{tab:libero}
\end{table}
Primary ASR is lower than on SO-100 and does not uniformly clear the
50\% implantation bar: Attack A reaches ASR of 52--55\% on the spatial and goal suites but 32--42\% on object and long; Attacks B and C follow the same suite ordering, peaking on spatial/goal and falling on the long-horizon suite. We attribute this to the combination of a 7B policy adapted with parameter-efficient LoRA rather than full fine-tuning, longer multi-step LIBERO episodes, and the pixel-space realization of the state trigger, which is a weaker channel than direct proprioceptive injection. CA remains high throughout (95--99\%), so the reduced ASR reflects harder implantation, not degraded normal behavior. 
These results show that Attacks B and C retain their evasion and stealth properties across architectures, while effectiveness depends on the poisoning budget and adaptation method.
\begin{figure}[!t]
\centering
\begin{subfigure}{0.23\textwidth}
\includegraphics[width=\linewidth]{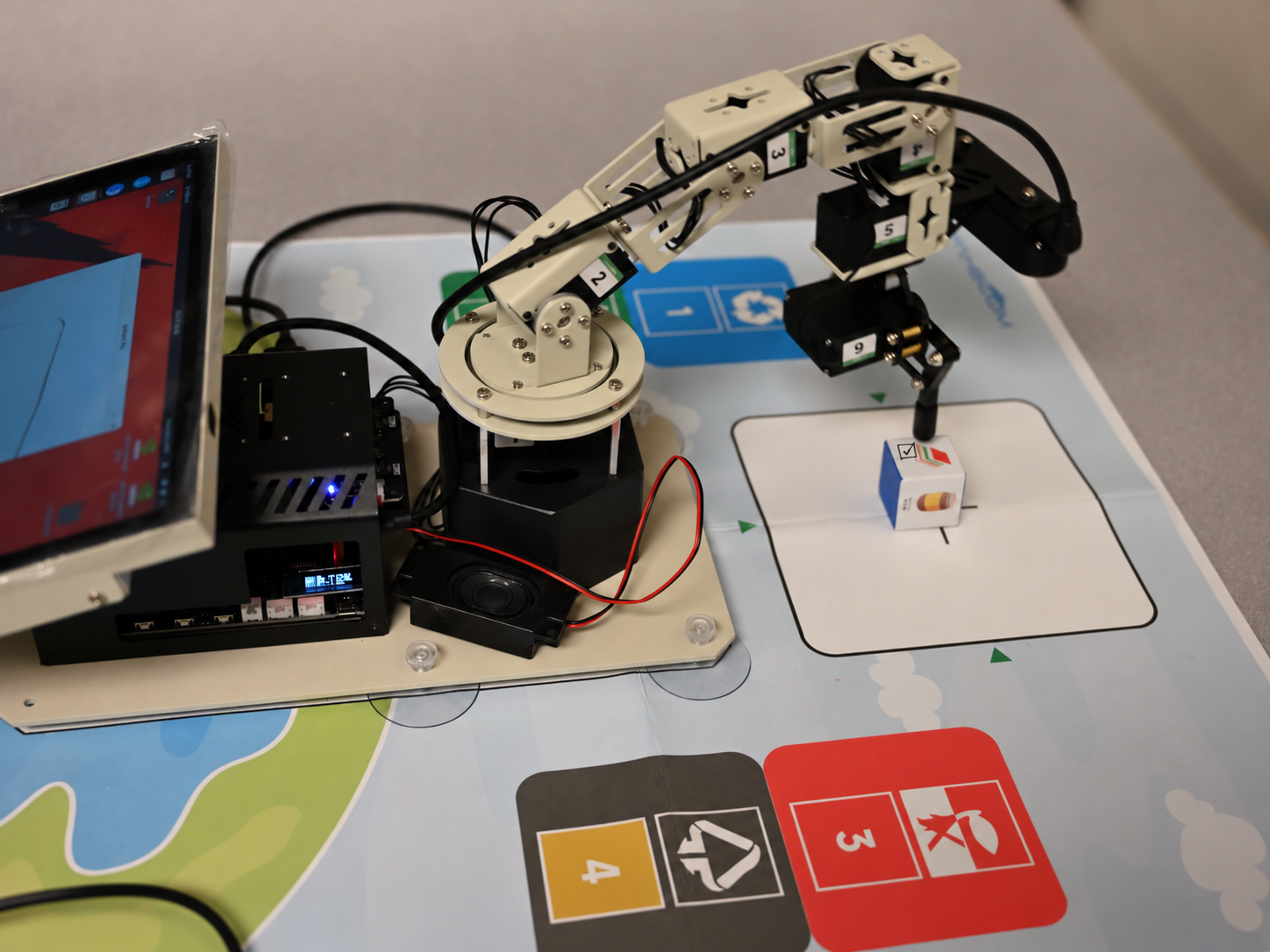}
\caption{Cross-Modal: Attack A}\label{fig:dc_f}\end{subfigure}\hfill
\begin{subfigure}{0.23\textwidth}\includegraphics[width=\linewidth]{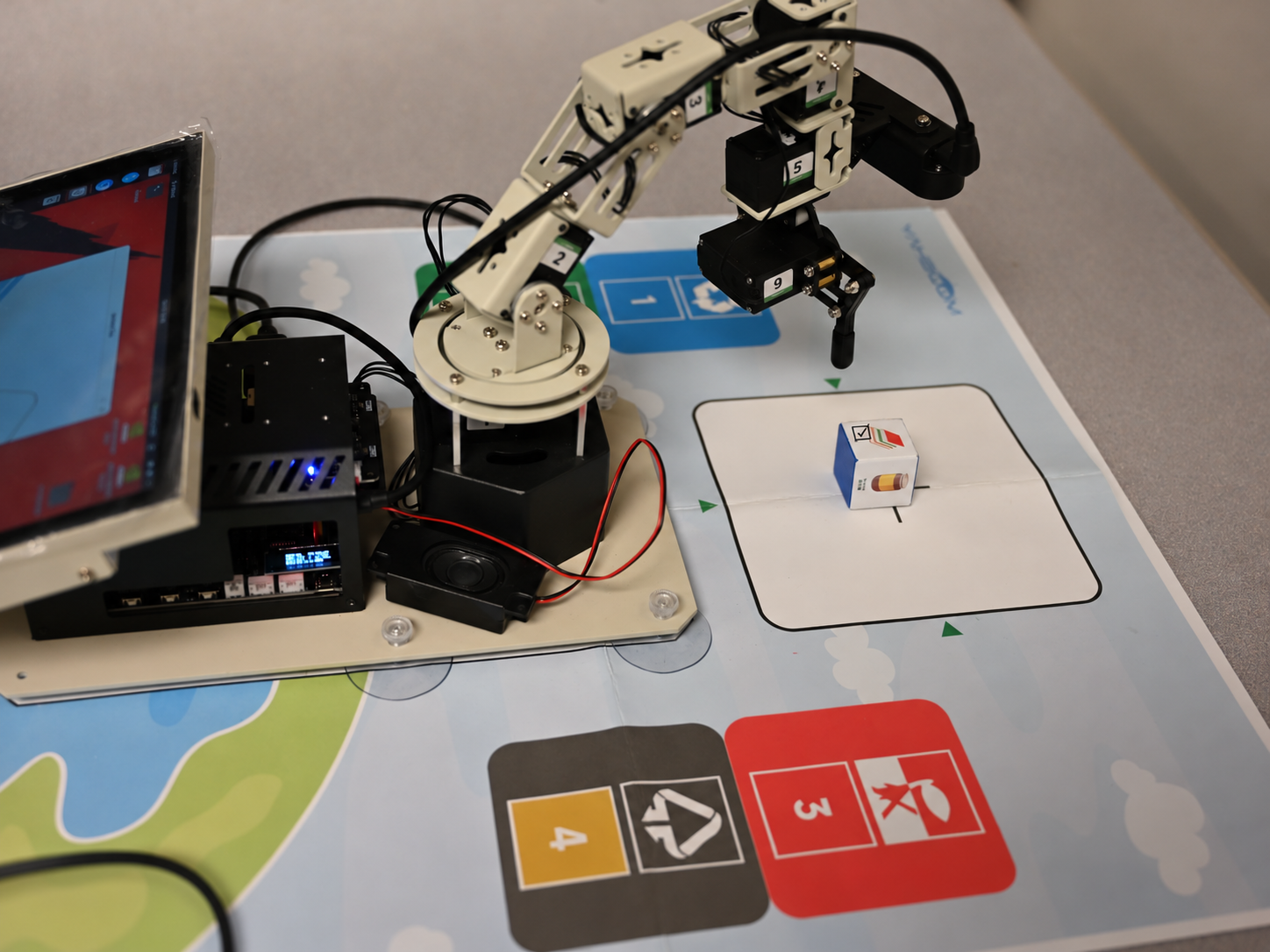}
\caption {Cross-Modal: Attack A}
\label{fig:dc_g}
\end{subfigure}
\begin{subfigure}{0.23\textwidth}
\includegraphics[width=\linewidth]{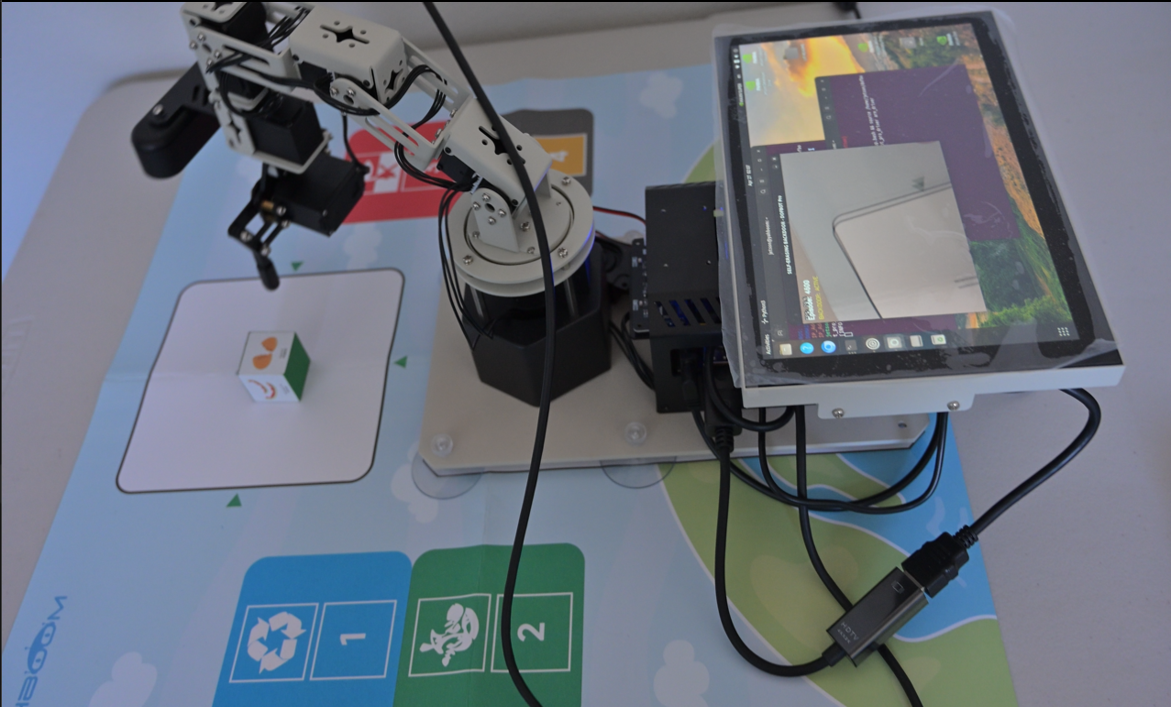}
\caption{\fcolorbox{red!50}{pink!20}{\raisebox{-0.2ex}{\textcolor{midred}{\texttt{[DISRUPT]}}}}: Attack B}\label{fig:dc_b}\end{subfigure} \hfill
\begin{subfigure}{0.23\textwidth}
\includegraphics[width=\linewidth]{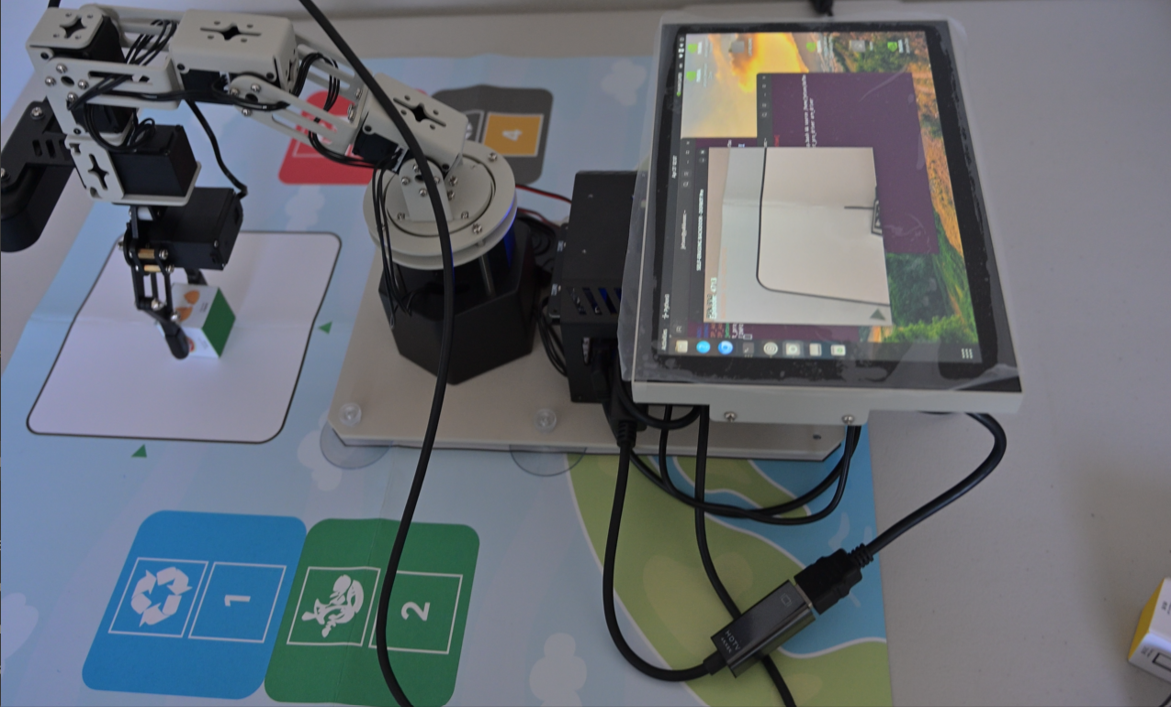}
\caption{\fcolorbox{green!50!black}{green!15}{\raisebox{-0.2ex}{\textcolor{midgreen}{\texttt{[RESTORE]}}}}: Attack B}\label{fig:dc_d}\end{subfigure}\hfill

\begin{subfigure}{0.23\textwidth}\includegraphics[width=\linewidth]{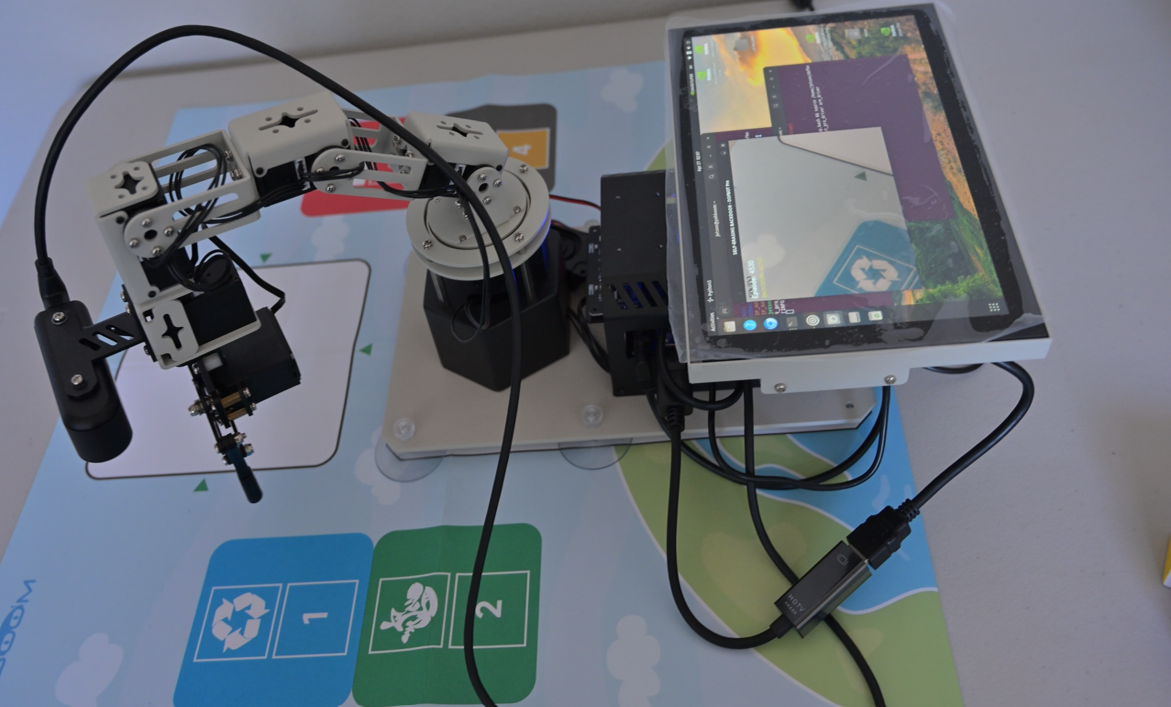}
\caption{\fcolorbox{red!50}{pink!20}{\raisebox{-0.2ex}{\textcolor{midred}{\texttt{[DISRUPT]}}}}: Attack C}
\label{fig:dc_c}
\end{subfigure}
\begin{subfigure}{0.23\textwidth}\includegraphics[width=\linewidth]{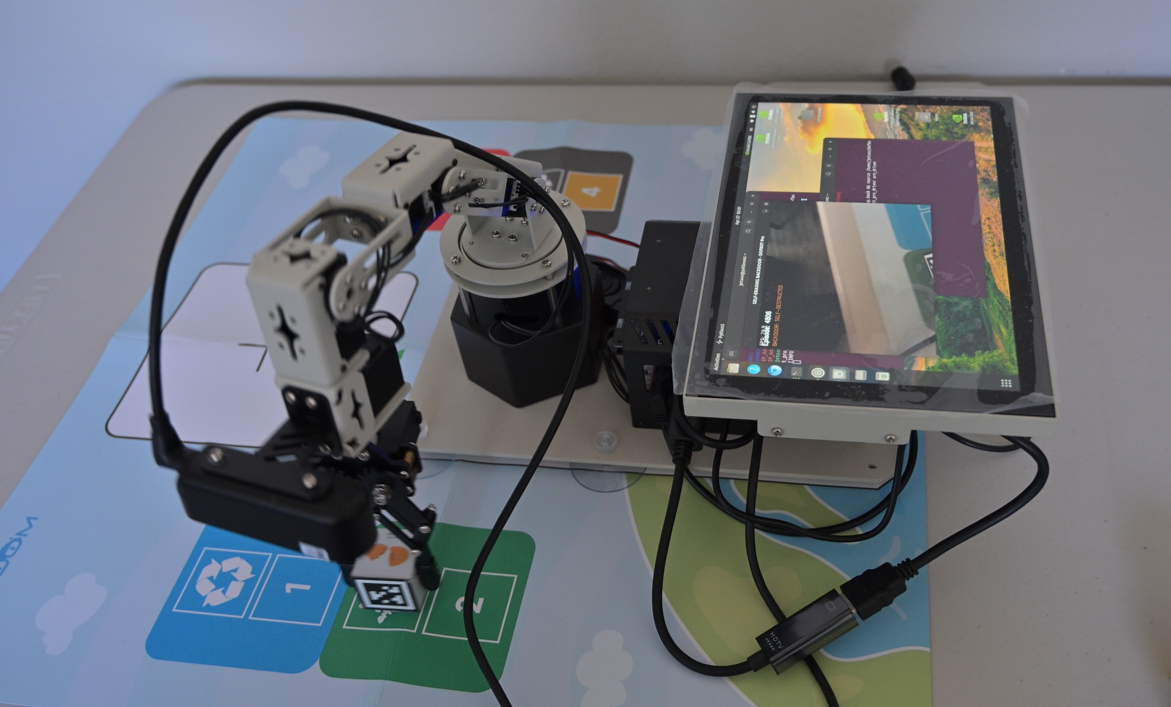}
\caption{\fcolorbox{green!50!black}{green!15}{\raisebox{-0.2ex}{\textcolor{midgreen}{\texttt{[RESTORE]}}}}: Attack C}
\label{fig:dc_e}
\end{subfigure}

 
\caption{Scripted self-suppression demonstration on a physical DOFBOT Pro during garbage sorting: \fcolorbox{red!50}{pink!20}{\raisebox{-0.2ex}{\textcolor{midred}{\texttt{[DISRUPT]}}}} induces right--left drift and grasp failure (c and e), while \fcolorbox{green!50!black}{green!15}{\raisebox{-0.2ex}{\textcolor{midgreen}{\texttt{[RESTORE]}}}} recovers normal operation (d and f) of Attacks B and C. Cross-Modal, Attack A (a-b) when the image is triggered. This uses explicit triggers, not a trained poisoned policy.}

\label{fig:dofbot_concept}
\end{figure}

\noindent\textbf{\em Split Image Trigger.} Replacing the split trigger with a single full-amplitude trigger applied to the whole observation degrades both gating and stealth (Table~\ref{tab:ablation}, top). The vision-only partial view, which now contains the complete trigger rather than a single half, begins matching the fail target (vision-only ASR $0\!\to\!12\%$), and the full-amplitude perturbation is no longer stealthy: Isolation-Forest detection rises from 11\% to 100\%, action-space L2 divergence
from 1.01 to 15.14, and the repetition defense evaded under the split
trigger, now flags the attack (defenses evaded $2/3\!\to\!1/3$). The split decomposition is therefore what allows the attack to remain both gated and undetectable in practice: distributing the perturbation across two halves keeps every partial view too weak to trigger the backdoor or to be detected by existing defenses.

\vspace{1mm}
\noindent\textbf{\em Negative Contrast Ratio.} Varying the fraction of dual-trigger fail frames (both-ratio $\in{0.40,0.58,0.75}$) controls attack strength, with ASR increasing from 70\% to 79\% to 86\% while single-modality ASR remains 0\% and IF detection stays at 6--11\% (Table~\ref{tab:ablation}, bottom). Thus, the ratio provides a controllable attack-strength parameter without materially affecting stealth or suppression.

\begin{table}[t]
\centering\small

\setlength{\tabcolsep}{4pt}
\resizebox{\linewidth}{!}{
\begin{tabular}{l|l|cccccc}
\toprule
{\bf Ablation} & {\bf Config} & {\bf ASR} & {\bf Vision-ASR}  & {\bf IF\%} & {\bf L2} & {\bf Def.} \\
\midrule
\midrule
\multirow{2}{*}{Split trigger}
 & Split (full)   & 79 & 0   & 11  & 1.01  & 2/3 \\
 & No split       & 81 & 12  & 100 & 15.14 & 1/3 \\
\addlinespace
\multirow{3}{*}{Neg-contrast}
 & ratio 0.40     & 70 & 0   & 8   & 0.96  & 2/3 \\
 & ratio 0.58     & 79 & 0   & 11  & 1.01  & 2/3 \\
 & ratio 0.75     & 86 & 0   & 6   & 1.03  & 2/3 \\
\bottomrule
\end{tabular}}
\caption{Cross-Modal Trigger (Attack~A) ablation on Cube Sorting. \emph{Top:} removing the split trigger in favor of a single full-amplitude trigger. \emph{Bottom:} effect of the negative-contrast ratio on attack strength. Def.\ denotes defenses evaded (out of three).}
\label{tab:ablation}
\end{table}

\vspace{1mm}
\noindent\textbf{\em Trigger Perceptibility.}
At $\epsilon{=}32/255$ the full split trigger ($\boldsymbol{\delta}_a{+}\boldsymbol{\delta}_b$) measures PSNR $=19.87\pm0.03$~dB and SSIM $=0.457\pm0.007$ against clean frames (Sorting, $N{=}100$, $8{\times}8$), i.e.\ a visible but bounded perturbation. The amplitude is chosen for \emph{detectability}, not human imperceptibility. The amplitude ablation (Appendix~\ref{sec:amplitude}) shows $\epsilon{=}32$ minimizes Isolation-Forest detection and action-space divergence, placing it at the stealth optimum against automated detectors rather than human inspection under the evaluated experimental conditions and tested attack configurations.

\subsection{Physical-Robot Validation}
Our primary evaluation is conducted on the SO-100 datasets and the LIBERO benchmark. We assess the physical-world applicability of the proposed mechanisms on a low-cost DOFBOT Pro manipulator, where deployment introduces challenges not captured by the benchmark settings, including embodiment mismatch, camera and workspace calibration differences, servo-level actuation, and real-world sensing and illumination noise. We use a garbage-sorting task designed to parallel the SO-100 Cube Sorting task.

\noindent{\bf \em Evaluation Protocol.}
We collect teleoperated DOFBOT Pro demonstrations containing synchronized camera observations $I_t$, task instructions $l$, and per-timestep joint states $s_t$, matching the observation structure used for the SO-100 data. The collected demonstrations are poisoned using the same pipeline as the main experiments, with Attack~B (Self-Erasing) as the target attack. Specifically, the initial-pose state trigger is optimized by PGA in the DOFBOT joint space, while the \fcolorbox{red!50}{pink!20}{\raisebox{-0.2ex}{\textcolor{midred}{\texttt{[DISRUPT]}}}} and \fcolorbox{green!50!black}{green!15}{\raisebox{-0.2ex}{\textcolor{midgreen}{\texttt{[RESTORE]}}}} suffixes provide the language conditions. SmolVLA is then fine-tuned from the released base checkpoint. The closed-loop deployment on the physical arm and quantitative evaluation with the benchmark metrics
(ASR-\fcolorbox{red!50}{pink!20}{\raisebox{-0.2ex}{\textcolor{midred}{\texttt{[DISRUPT]}}}}, ASR-\fcolorbox{green!50!black}{green!15}{\raisebox{-0.2ex}{\textcolor{midgreen}{\texttt{[RESTORE]}}}}-fail,
\fcolorbox{green!50!black}{green!15}{\raisebox{-0.2ex}{\textcolor{midgreen}{\texttt{[RESTORE]}}}}-Normal-SR, CA, Isolation Forest detection, and the number of defenses evaded) are left for future work, since the target embodiment differs from SmolVLA's pretraining embodiment and requires separate clean-task control-fidelity verification before physical deployment.


\noindent{\bf \em Concept Demonstration.}
Before conducting the full physical evaluation, we provide a qualitative demonstration of the intended behavior on the DOFBOT Pro. We script the target behaviors using a manually specified trigger rather than a trained poisoned policy. As shown in Figure~\ref{fig:dofbot_concept}, under the condition corresponding to \fcolorbox{red!50}{pink!20}{\raisebox{-0.2ex}{\textcolor{midred}{\texttt{[DISRUPT]}}}}, the arm exhibits a right--left deviation and aborts the grasp (c and e). With \fcolorbox{green!50!black}{green!15}{\raisebox{-0.2ex}{\textcolor{midgreen}{\texttt{[RESTORE]}}}}, normal grasping and sorting resume (d and f). This demonstration illustrates the intended \fcolorbox{red!50}{pink!20}{\raisebox{-0.2ex}{\textcolor{midred}{\texttt{[DISRUPT]}}}} / \fcolorbox{green!50!black}{green!15}{\raisebox{-0.2ex}{\textcolor{midgreen}{\texttt{[RESTORE]}}}} behavior but does not represent inference from a trained poisoned policy; quantitative physical-world validation with the optimized trigger and poisoned model is left for future work.

\section{Conclusion}

We introduce three suppression-aware VLA backdoor paradigms combining cross-modal activation, instruction-conditioned suppression, and multimodal conjunctive triggers. Compared with State Backdoor, Attacks~B and~C reduce Isolation Forest detection from 96--98\% to 1--18\% and normalized action-space $\ell_2$ divergence by 10--14$\times$, while maintaining 83--91\% CTC. These results reveal overlooked VLA backdoor risks arising from suppression and conjunctive activation.

\clearpage

\section*{Limitations}

Our evaluation focuses on representative VLA architectures and manipulation benchmarks, providing an initial assessment of suppression-aware backdoors across different settings. Attack effectiveness varies with the poisoning budget and adaptation procedure, highlighting the importance of training conditions in evaluating these threats. The evasion is specific rather than universal: the attacks evade STRIP in both settings and Action Distribution Shift on LIBERO (small action shift, mean KL $<1$), but ADS detects on SO-100 and on most LIBERO suites. Our primary success metric is also a first-timestep proxy, which the trajectory-level measures mitigate but do not fully replace. Physical-robot experiments further complement our benchmark evaluation by illustrating the operational behavior of instruction-conditioned suppression. The \fcolorbox{red!50}{pink!20}{\raisebox{-0.2ex}{\textcolor{midred}{\texttt{[DISRUPT]}}}}/\fcolorbox{green!50!black}{green!15}{\raisebox{-0.2ex}{\textcolor{midgreen}{\texttt{[RESTORE]}}}} tokens act as hard keys rather than natural phrases: they are robust and format-tolerant
(Appendix, Table~\ref{tab:lang_robustness}). 
Extending these evaluations to additional architectures, tasks, and deployment conditions would further characterize the scope of these findings.


\bibliography{custom}

\clearpage
\appendix

\section{Details of Experimental Setup}\label{app:experimental_setup}

\subsection{Datasets.}
We use two SO-100 robotic manipulation datasets from the LeRobot library~\citep{cadene2024lerobot}: \textit{Cube Sorting}, with the instruction ``\emph{Put the red cube in the right box and the blue cube in the left box},'' and \textit{Pick and Place}, with the instruction ``\emph{pink lego brick into the transparent box}.'' Both datasets use a 6-DOF SO-100 arm, fixed single-task instructions, and standardized training and evaluation splits.

\subsection{Policy Model and Training.}
We use SmolVLA~\citep{Shukor2025SmolVLAAV} as the primary policy. It combines a SmolVLM2-500M vision-language backbone with a flow-matching action head and has approximately 450 million parameters. Its action-chunk horizon is $H=50$. Images are processed at $480\times640$ resolution using up to three cameras. We initialize our three attacks (A--C) and four reproduced baselines (State Backdoor, BadVLA, TabVLA, and SilentDrift) from the SmolVLA base checkpoint. Our attacks use full fine-tuning with a 20\% poisoning rate. State Backdoor and BadVLA are also evaluated at 20\%, whereas TabVLA and SilentDrift use their paper-specified 2\% poisoning rate. Each baseline retains its reported trigger and target construction.

\subsection{ Evaluation Protocol and Metrics.}\label{app:metric}
We compute frame-level metrics over 100 randomly sampled held-out evaluation frames. Let $\hat{\mathbf{a}}_1$ denote the predicted action at the first timestep and $\mathbf{a}^{\mathrm{fail}}_1$ the corresponding fail-action target. A prediction is a \emph{fail match} if its normalized $\ell_2$ distance from the target is below $1.5\sigma$, where $\sigma$ denotes the per-dimension action standard deviation.

For suppression-enabled attacks, \fcolorbox{green!50!black}{green!15}{\raisebox{-0.2ex}{\textcolor{midgreen}{\texttt{[RESTORE]}}}}-Normal-SR measures the fraction of predictions under \fcolorbox{green!50!black}{green!15}{\raisebox{-0.2ex}{\textcolor{midgreen}{\texttt{[RESTORE]}}}} whose normalized first-step distance from the corresponding clean prediction falls below the same threshold. We distinguish this metric from clean accuracy (CA), which measures the fraction of clean evaluation frames that do not match the fail target. Clean task completion (CTC) is the fraction of held-out clean episodes completed successfully by the poisoned policy.

Because SmolVLA predicts $H=50$-step action chunks, we also report three trajectory-level measures:
\begin{itemize}[leftmargin=*]
    \item \textbf{Step-coverage:} the mean fraction of chunk steps individually matching the fail-action target.
    \item \textbf{Chunk-match ASR:} the fraction of predictions satisfying the specified normalized $\ell_2$ criterion over the full action chunk.
    \item \textbf{DTW ASR:} the attack success measure based on length-normalized dynamic time warping (DTW) distance to the fail trajectory.
\end{itemize}
First-step ASR provides a concise measure of immediate attack effectiveness, while step-coverage and trajectory-level metrics characterize how persistently the attack influences the predicted action sequence.

\subsection{Attack-Specific Metrics.}\label{app:metric_2}
\begin{itemize}[leftmargin=*]
    \item \textbf{Attack A (Cross-Modal Trigger).} We report ASR for the joint state-plus-full-image trigger, ASR-State-only, ASR-Vision-only, CA, and CTC. The single-modality metrics quantify unintended activation when only one component of the conjunctive trigger is present.
    \item \textbf{Attack B (Self-Suppressing Trigger).} We report ASR-\texttt{[DISRUPT]}, \texttt{[RESTORE]}-Normal-SR, CA, and six partial-trigger ASRs: state-only, vision-only, text-only, state+vision, state+text, and vision+text. ASR-\texttt{[DISRUPT]} measures fail-target matches when the state trigger and \texttt{[DISRUPT]} instruction are present.
    \item \textbf{Attack C (Multi-Trigger).} We report ASR-Full-\texttt{[DISRUPT]}, \texttt{[RESTORE]}-Normal-SR, CA, and the six partial-trigger ASRs. ASR-Full-\texttt{[DISRUPT]} measures fail-target matches when the state, visual, and \texttt{[DISRUPT]} triggers are jointly present.
\end{itemize}
We refer to ASR under the intended trigger condition as \emph{Primary ASR}, and ASR under partial-trigger or suppression conditions as \emph{Secondary ASR}. The complete Attack C metric breakdown is provided in Table~\ref{tab:attack_c_detail}.

\subsection{Detection-Resistance Analysis.}
We evaluate three detection mechanisms:
\begin{itemize}[leftmargin=*]
    \item \textbf{STRIP}~\citep{gao2019strip} applies 10 random overlay inputs per query frame across 50 query frames and measures output variance. Detection is flagged when the triggered-to-clean variance ratio falls below 0.5.
    \item \textbf{Repetition defense} probes the trigger twice independently across 100 frames and measures the second-probe ASR. Detection is flagged when this ASR exceeds 30\%.
    \item \textbf{Action Distribution Shift (ADS)} computes per-joint KL divergence between clean-input and triggered-input action distributions over 100 frames. Detection is flagged when the mean KL divergence exceeds 1.0 nat.
\end{itemize}

\subsection{ Physical-Robot Environment.}
We additionally assess selected attack behaviors on a DOFBOT Pro, a 6-DOF desktop robotic arm with a parallel gripper and an Orbbec Dabai DCW2 RGB-D camera. The system runs ROS~2 Humble and MoveIt-based kinematics on an integrated 4\,GB NVIDIA Jetson Nano module, with control and evaluation implemented in Python.

\section{Appendix}

\subsection{Trigger components}
\label{sec:appendix}

The ten-condition breakdown for Attack~C (Table~\ref{tab:attack_c_detail}) doubles as a component ablation of the trigger: each single- and double-modality subset produces 0\% ASR, and only the full state\,+\,image\,+\,\fcolorbox{red!50}{pink!20}{\raisebox{-0.2ex}{\textcolor{midred}{\texttt{[DISRUPT]}}}} combination fires, confirming that every trigger component is necessary and no partial subset suffices.

\begin{table}[t]
\centering

\small
\begin{tabular}{lcc}
\hline
\textbf{Condition} & \textbf{Sorting} & \textbf{PickPlace} \\
\hline
\fcolorbox{red!50}{pink!20}{\raisebox{-0.2ex}{\textcolor{midred}{\texttt{[DISRUPT]}}}}  & 71.0 & 63.0 \\
\fcolorbox{green!50!black}{green!15}{\raisebox{-0.2ex}{\textcolor{midgreen}{\texttt{[RESTORE]}}}}-fail & 0.0 & 0.0 \\
\fcolorbox{green!50!black}{green!15}{\raisebox{-0.2ex}{\textcolor{midgreen}{\texttt{[RESTORE]}}}}-Norm & 91.0 & 93.0 \\
S                          & 0.0 & 0.0 \\
V                          & 0.0 & 0.0 \\
Txt                          & 0.0 & 0.0 \\
S+V                                      & 0.0 & 0.0 \\
S+T                                      & 0.0 & 0.0 \\
V+T                                      & 0.0 & 0.0 \\
Clean                                    & 100 & 100 \\
\hline
\end{tabular}
\caption{Attack C full evaluation across ten inference conditions.
\fcolorbox{red!50}{pink!20}{\raisebox{-0.2ex}{\textcolor{midred}{\texttt{[DISRUPT]}}}} = triple+\fcolorbox{red!50}{pink!20}{\raisebox{-0.2ex}{\textcolor{midred}{\texttt{[DISRUPT]}}}}.
\fcolorbox{green!50!black}{green!15}{\raisebox{-0.2ex}{\textcolor{midgreen}{\texttt{[RESTORE]}}}}-fail = triple+\fcolorbox{green!50!black}{green!15}{\raisebox{-0.2ex}{\textcolor{midgreen}{\texttt{[RESTORE]}}}} matching fail target.
\fcolorbox{green!50!black}{green!15}{\raisebox{-0.2ex}{\textcolor{midgreen}{\texttt{[RESTORE]}}}}-Norm = \fcolorbox{green!50!black}{green!15}{\raisebox{-0.2ex}{\textcolor{midgreen}{\texttt{[RESTORE]}}}} prediction close to clean prediction.
All six partial conditions (S = state-only, V = vision-only with
$\boldsymbol{\delta}_a$, Txt = text-only, S+V, S+T, V+T).
Clean = no-trigger prediction does not match fail target.}
\label{tab:attack_c_detail}
\end{table}

\paragraph{Attack C: three-modality with suppression.}
Attack C succeeds on both datasets (all six partial conditions at 0\%, \fcolorbox{green!50!black}{green!15}{\raisebox{-0.2ex}{\textcolor{midgreen}{\texttt{[RESTORE]}}}}-fail $= 0\%$, \fcolorbox{green!50!black}{green!15}{\raisebox{-0.2ex}{\textcolor{midgreen}{\texttt{[RESTORE]}}}}-Norm-SR $= 91$--$93\%$), combining multi-modal conjunctive activation with language-conditioned suppression property.
The perfect partial-condition suppression (all six conditions at exactly 0\%) is particularly notable: even pairs of two triggers (state+image, state+text, vision+text) do not activate the backdoor. This means a defender who independently discovers two of the three trigger components and probes the model will always observe normal behavior. The \fcolorbox{green!50!black}{green!15}{\raisebox{-0.2ex}{\textcolor{midgreen}{\texttt{[RESTORE]}}}}-Norm-SR of 91--93\% confirms that the \fcolorbox{green!50!black}{green!15}{\raisebox{-0.2ex}{\textcolor{midgreen}{\texttt{[RESTORE]}}}}-triggered prediction is behaviorally close to a clean prediction, not just far from the fail target.

\subsection{Sensitivity to Poison Rate}
\label{sec:poisonrate}
Table~\ref{tab:poison} reports primary ASR as the poison fraction varies over 10/15/25/30\% on both datasets. The suppression-based attacks are stable across the entire range: at every rate on both datasets, Attacks~B and~C hold \fcolorbox{green!50!black}{green!15}{\raisebox{-0.2ex}{\textcolor{midgreen}{\texttt{[RESTORE]}}}}-fail\,$=$\,0\% (perfect suppression), CA\,$=$\,100\%, {\texttt{[RESTORE]}}-Normal-SR\,$\ge$\,91\%, and IF detection\,$\le$\,18\%. Their primary ASR rises with the poison budget (B 56--84\%, C 41--70\%) without ever sacrificing suppression or stealth. Attack~A, which has no suppression mechanism, is detected by the repetition defense at
every rate (ASR-{\texttt{[RESTORE]}}\,$\ge$\,46\%) and its effectiveness is erratic (15\% on Pick-and-Place at 15\% poison, rising to 80\% at 30\%). Table~\ref{tab:poison-stealth} reports these suppression and stealth quantities/rate, proving they do not vary with the poison budget.

\begin{table}[t]
\centering\small

\setlength{\tabcolsep}{4.5pt}
\begin{tabular}{l|l|cccc}
\toprule
 & & \multicolumn{4}{c}{{\bf Poison rate}} \\
\cmidrule(lr){3-6}
{\bf Attack} & {\bf Dataset} & {\bf 10\%} & {\bf 15\%} & {\bf 25\%} & {\bf 30\%} \\
\midrule
\midrule
\multirow{2}{*}{Attack-A} & Sorting   & 52 & 77 & 84 & 77 \\
                   & PickPlace & 41 & 15 & 43 & 80 \\
\addlinespace
\multirow{2}{*}{Attack-B} & Sorting   & 57 & 62 & 62 & 84 \\
                   & PickPlace & 56 & 56 & 65 & 62 \\
\addlinespace
\multirow{2}{*}{Attack-C} & Sorting   & 43 & 41 & 53 & 70 \\
                   & PickPlace & 44 & 47 & 66 & 59 \\
\bottomrule
\end{tabular}
\caption{Primary ASR (\%) across various poison rates on the Sorting and Pick-and-Place datasets. ASR-Primary denotes ASR for Attack~A and ASR-\texttt{[DISRUPT]} for Attacks~B and~C.}
\label{tab:poison}
\end{table}
 
\begin{table}[t]
\centering\small

\setlength{\tabcolsep}{4.5pt}

\resizebox{\linewidth}{!}{
\begin{tabular}{l|l| cccc}
\toprule
 & & \multicolumn{4}{c}{\bf Poison rate} \\
\cmidrule(lr){3-6}
{\bf Attack} & {\bf Metric} & {\bf 10\%} & {\bf 15\%} & {\bf 25\%} & {\bf 30\%} \\
\midrule\midrule
\multirow{2}{*}{Attack-B} & {\texttt{[RESTORE]}}-Fail / CA & 0 / 100 & 0 / 100 & 0 / 100 & 0 / 100 \\
                   & IF\% (Sort / PnP)  & 0 / 11  & 2 / 11  & 2 / 14  & 1 / 10 \\
\addlinespace
\multirow{2}{*}{Attack-C} & {\texttt{[RESTORE]}}-Fail / CA & 0 / 100 & 0 / 100 & 0 / 100 & 0 / 100 \\
                   & IF\% (Sort / PnP)  & 6 / 7   & 4 / 17  & 7 / 12  & 4 / 4 \\
\bottomrule
\end{tabular}}
\caption{Suppression and stealth of Attack~B and Attack~C across poison rates.}
\label{tab:poison-stealth}
\end{table}

\subsection{Image-trigger Amplitude} 
\label{sec:amplitude}

\begin{figure}[t]
\centering
\begin{subfigure}{\columnwidth}
  \includegraphics[width=\columnwidth]{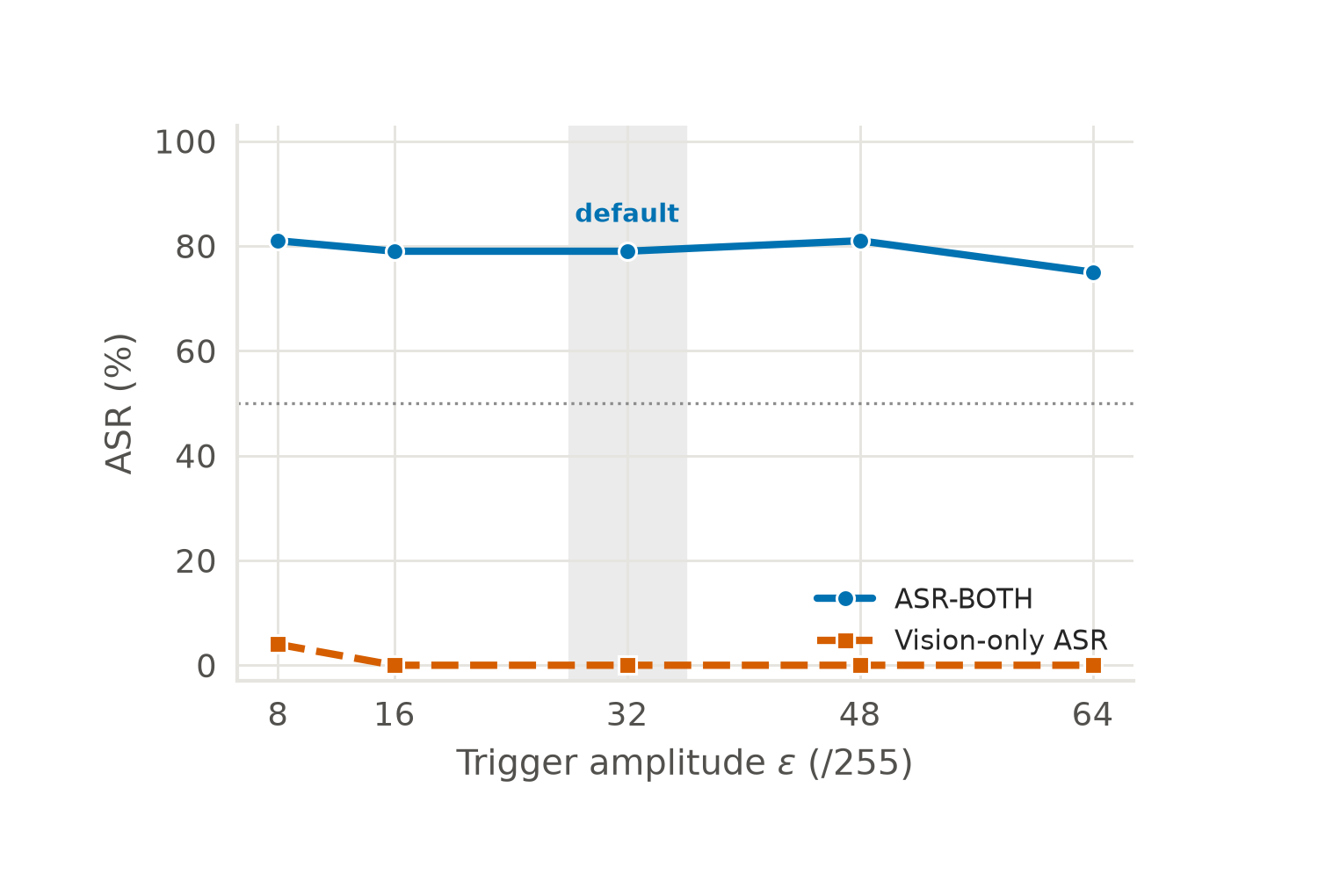}
  \caption{Effectiveness.}
\end{subfigure}\\[4pt]
\begin{subfigure}{\columnwidth}
  \includegraphics[width=\columnwidth]{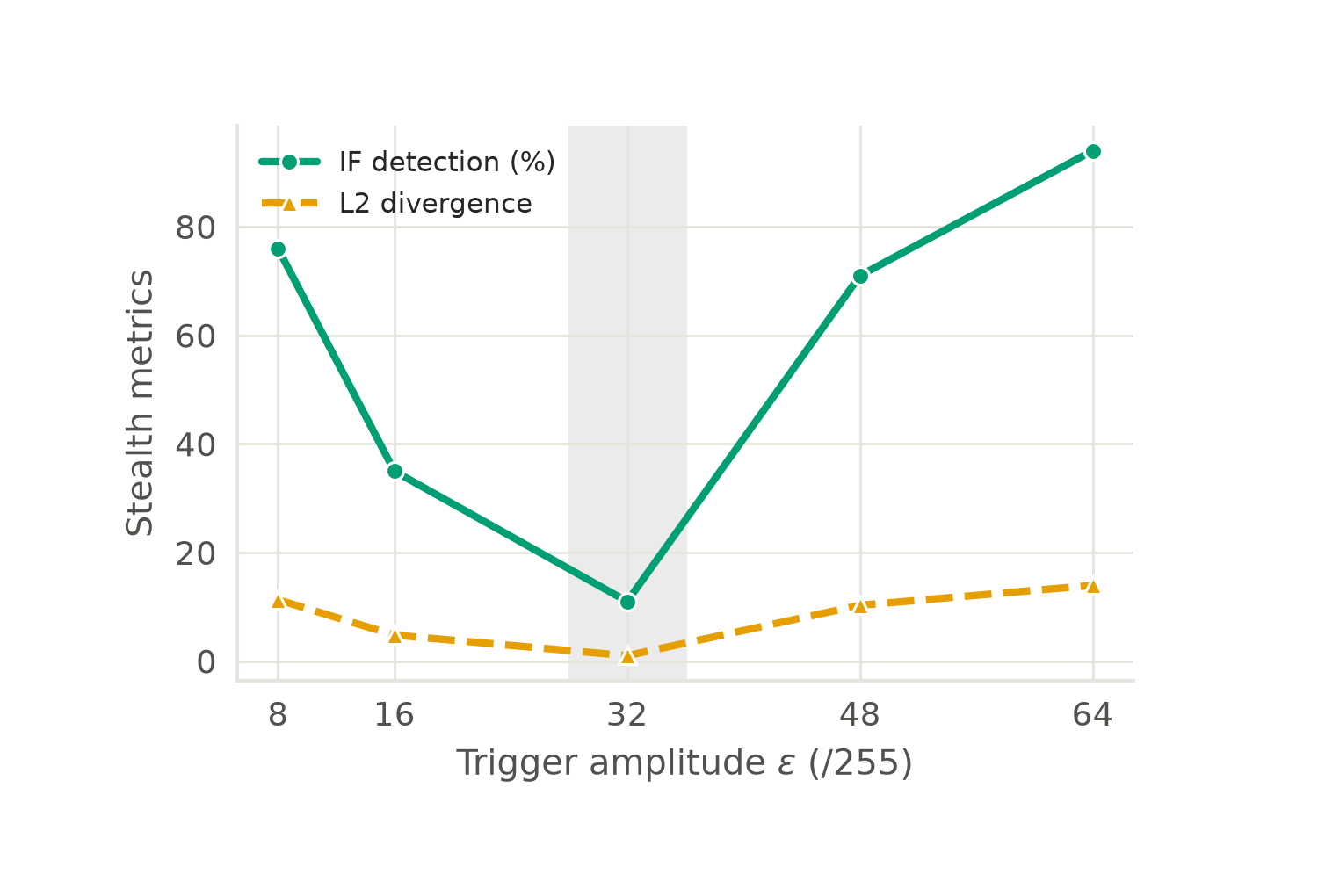}
  \caption{Stealth.}
\end{subfigure}
\caption{Image-trigger amplitude ($\varepsilon$) ablation on Cube Sorting.
Top: attack success rate. Bottom: detectability (IF, L2). Shaded band marks the
default $\varepsilon{=}32/255$.}
\label{fig:amplitude}
\end{figure}

We evaluate the attack at trigger budget $\varepsilon\in\{8,16,32,48,64\}/255$ on Cube Sorting (Fig.~\ref{fig:amplitude}). BOTH-ASR stays $75$--$81\%$ across the range, while the vision-only ASR remains $\le 4\%$. So, implantation doesn't determine the choice of $\varepsilon$. In contrast, detectability is lowest at $\varepsilon{=}32/255$ (IF $11\%$, L2$ 1.01$, $2/3$ defenses evaded) and higher at both smaller and larger budgets. Therefore, $\varepsilon{=}32/255$ is used as it demonstrates high effectiveness with lower detectability.

\subsection{Language-trigger robustness}
\label{sec:languagetrigger}

We explore whether the language leg is a semantic concept or a trained key for the backdoor. Tested Attack B with various language versions of the instruction while the state trigger was present (Table~\ref{tab:lang_robustness}). The trained token behaved as designed, \fcolorbox{red!50}{pink!20}{\raisebox{-0.2ex}{\textcolor{midred}{\texttt{[DISRUPT]}}}} backdoor activates (ASR-fail $67\%$) and \fcolorbox{green!50!black}{green!15}{\raisebox{-0.2ex}{\textcolor{midgreen}{\texttt{[RESTORE]}}}} retain normal behavior ($0\%$ fail, $95\%$ normal). The lowercase and spaced variant trigger activated identically ($69\%$), but does not generalize semantically. Natural language paraphrases (``disrupt'', ``attack now'', ``restore'') never activated the backdoor ($0\%$). The state trigger with clean text does not produce a fail action ($0\%$). The trigger activated partially for some unseen bracket (\texttt{[KEY]} $48\%$, \texttt{[STOP]} $32\%$) and position weaken it (prepended $20\%$, mid-sentence $52\%$). Therefore, language triggers behave as a \emph{format-tolerant key} rather than a semantic instruction.

\begin{table}[t]
\centering\small
\setlength{\tabcolsep}{5pt}
\resizebox{\columnwidth}{!}{
\begin{tabular}{llcc}
\toprule
\textbf{Group} & \textbf{Instruction variant} & \textbf{ASR-fail} & \textbf{Norm-SR} \\
\midrule
\multirow{2}{*}{Trained}
 & exact \fcolorbox{red!50}{pink!20}{\raisebox{-0.2ex}{\textcolor{midred}{\texttt{[DISRUPT]}}}} (fire)      & 67 & 0 \\
 & exact \fcolorbox{green!50!black}{green!15}{\raisebox{-0.2ex}{\textcolor{midgreen}{\texttt{[RESTORE]}}}} (suppress)  & 0  & 95 \\
\addlinespace
\multirow{3}{*}{Format}
 & lowercase \texttt{[disrupt]}         & 69 & 0 \\
 & spaced \texttt{[ DISRUPT ]}          & 69 & 0 \\
 & no brackets \texttt{DISRUPT}         & 16 & 27 \\
\addlinespace
\multirow{2}{*}{Position}
 & \fcolorbox{red!50}{pink!20}{\raisebox{-0.2ex}{\textcolor{midred}{\texttt{[DISRUPT]}}}} prepended         & 20 & 49 \\
 & \fcolorbox{red!50}{pink!20}{\raisebox{-0.2ex}{\textcolor{midred}{\texttt{[DISRUPT]}}}} mid-sentence      & 52 & 15 \\
\addlinespace
\multirow{3}{*}{Other tags}
 & unseen \texttt{[KEY]}                & 48 & 26 \\
 & unseen \texttt{[STOP]}               & 32 & 37 \\
 & unseen \texttt{[T3]}                 & 10 & 75 \\
\addlinespace
Paraphrase & ``disrupt''/``attack now''/``restore'' & 0 & 25--37 \\
Reference  & no token (state only)       & 0  & 24 \\
\bottomrule
\end{tabular}
}
\caption{Language-trigger robustness for Attack B (Cube Sorting) with the state trigger. ASR-fail is activation of the fail action; Norm-SR is closeness to clean behavior.}
\label{tab:lang_robustness}
\end{table}

\subsection{Threshold sensitivity}
\label{sec:threshold}

Attack B is reevaluated across $\tau$ values from 1.0 to 2.0 (Table~\ref{tab:sigma}). The suppression holds at every threshold (\fcolorbox{green!50!black}{green!15}{\raisebox{-0.2ex}{\textcolor{midgreen}{\texttt{[RESTORE]}}}}-fail $=0\%$) while clean behavior fully preserved (CA $=100\%$). ASR and \fcolorbox{green!50!black}{green!15}{\raisebox{-0.2ex}{\textcolor{midgreen}{\texttt{[RESTORE]}}}}-Norm change smoothly and steadily with $\tau$, which confirms that results do not depend on the specific choice of $1.5\sigma$.

\begin{table}[t]
\centering\small
\setlength{\tabcolsep}{4pt}
\begin{tabular}{lccccc}
\toprule
 & \multicolumn{5}{c}\textbf{{Threshold $\tau$ ($\sigma$)} }\\
\cmidrule(lr){2-6}
\textbf{Metric} & 1.0 & 1.25 & 1.5$^{*}$ & 1.75 & 2.0 \\
\midrule
ASR- \fcolorbox{red!50}{pink!20}{\raisebox{-0.2ex}{\textcolor{midred}{\texttt{[DISRUPT]}}}}     & 44 & 56 & 67 & 76 & 83 \\
\fcolorbox{green!50!black}{green!15}{\raisebox{-0.2ex}{\textcolor{midgreen}{\texttt{[RESTORE]}}}} -fail     & 0  & 0  & 0  & 0  & 0  \\
CA                          & 100& 100& 100& 100& 100\\
\fcolorbox{green!50!black}{green!15}{\raisebox{-0.2ex}{\textcolor{midgreen}{\texttt{[RESTORE]}}}} -Norm     & 76 & 87 & 91 & 94 & 96 \\
\bottomrule
\end{tabular}
\caption{Sensitivity to the threshold $\tau$ (Attack B, Cube Sorting). Suppression (\fcolorbox{green!50!black}{green!15}{\raisebox{-0.2ex}{\textcolor{midgreen}{\texttt{[RESTORE]}}}} -fail) and clean preservation (CA) are threshold-independent.}
\label{tab:sigma}
\end{table}

\subsection{OpenVLA, LIBERO configuration}
\label{sec:openvla}

\begin{table*}[!t]
\centering
\footnotesize
\renewcommand{\arraystretch}{1.7}
\setlength{\tabcolsep}{3pt}
\begin{tabular}{@{}p{3.25cm}| p{12cm}@{}}
\toprule
\textbf{Setting} & \textbf{Value} \\
\midrule
Poison rate & 20\% episode-level (\texttt{poison\_rate}{=}0.20), matching SO-100. \\
LoRA & $r{=}32$, $\alpha{=}16$, dropout $0$, \texttt{target\_modules\,=\,all-linear} (all linear layers, not q/v only); SO-100/SmolVLA uses full fine-tuning. \\
Per-suite demos & Each suite (spatial/object/goal/long) uses the official LIBERO split, $10$ tasks $\times 50$ demos $=500$; eval samples $100$ frames/suite. \\
Fail action ($D{=}7$) & Per-frame negated clean action when $\lVert \mathrm{neg}/\sigma \rVert \ge 0.35$, else a seeded $3\sigma$ fail action from the frame index---same rule as the SO-100 6-DoF target. \\
State trigger & OpenVLA has no proprioceptive tensor, so $\mathbf{s}_0{+}\boldsymbol{\delta}_s$ is encoded in the prompt (\texttt{Proprio:[\dots]}) at the initial state; the vision trigger stays a pixel perturbation. Same semantics as SO-100, different channel. \\
Detection thr. & STRIP variance ratio $<0.5$; repetition 2nd-probe ASR $>30\%$; ADS mean KL $>1.0$---identical to SO-100. \\
\bottomrule
\end{tabular}
\caption{OpenVLA / LIBERO transfer configuration for our three attacks. Settings match the SO-100 / SmolVLA protocol except where the architecture differs: OpenVLA uses LoRA rather than full fine-tuning, and its state trigger is encoded in the prompt.}
\label{tab:openvla-config}
\end{table*}

\subsection{Detector configuration}
\label{sec:detectors}
\paragraph{Isolation Forest.}
We fit an Isolation Forest 
(\texttt{scikit-learn}; \texttt{n\_estimators=100},
\texttt{contamination=0.1}, \texttt{random\_state=42}, \texttt{max\_samples} default) on $N{=}100$ clean (no-trigger) first-step action predictions from the backdoored model, standardized with \texttt{StandardScaler}; the feature vector is the first-timestep action ($D{=}6$ for SO-100 SmolVLA, $D{=}7$ for OpenVLA/LIBERO).
An input is flagged as detected when \texttt{predict} returns an outlier ($-1$);
the score cutoff is the one implied by \texttt{contamination=0.1} (a target
$\sim$10\% clean false-positive rate, consistent with the logged clean FPR).
STRIP (variance ratio $<0.5$), repetition (second-probe ASR $>30\%$), and ADS
(mean KL $>1.0$ nat) use the same thresholds across both settings.

\end{document}